\documentclass[]{fairmeta}
\usepackage{adjustbox}
\usepackage{array}      % if not already in the template
\usepackage{ragged2e}   % optional but nice for raggedright

\newcolumntype{L}[1]{>{\RaggedRight\arraybackslash}p{#1}}

\newcommand{\shortmmrb}{MMRB2}
\newcommand{\longmmrb}{Multimodal RewardBench 2}

\usepackage[most]{tcolorbox}
\tcbuselibrary{listings}

\newtcblisting{promptbox}{
  breakable,
  listing only,
  title=PROMPT,
  fonttitle=\bfseries\color{black}, % white title text
  coltitle=white,
  colbacktitle=white,               % black bar behind title
  colframe=black,                   % black border
  colback=white,                    % white interior
  boxrule=0.8pt,
  listing options={
    basicstyle=\ttfamily\small,
    breaklines=true,
    columns=fullflexible,
    numbers=none,
    backgroundcolor=\color{white}   % ensure no background in listings
  }
}

\lstdefinestyle{mypython}{
  language=Python,
  basicstyle=\ttfamily\scriptsize, % smaller text
  keywordstyle=\color{blue},
  stringstyle=\color{teal!60!black},
  commentstyle=\color{gray!60},
  numberstyle=\tiny\color{gray},
  numbers=left,
  stepnumber=1,
  numbersep=8pt,
  breaklines=true,
  tabsize=4,
  showstringspaces=false,
  backgroundcolor=\color{white},    % ensures no shading behind code
  xleftmargin=2pt,
  xrightmargin=2pt
}

\newtcblisting{pythonbox}{
  breakable,
  listing only,
  colback=white,         % white interior
  colframe=black,        % thin black border
  boxrule=0.4pt,
  listing options={style=mypython}
}

\title{Multimodal RewardBench 2: Evaluating Omni Reward Models for Interleaved Text and Image}

\author[*]{Yushi Hu}
\author[*]{Reyhane Askari-Hemmat}
\author[]{Melissa Hall}
\author[]{Emily Dinan}
\author[]{Luke Zettlemoyer}
\author[]{Marjan Ghazvininejad}

\affiliation[]{FAIR at Meta Superintelligence Labs}

\contribution[*]{Equal Contribution}

\abstract{
Reward models (RMs) are essential for training large language models (LLMs), but remain underexplored for omni models that handle interleaved image and text sequences. We introduce \textbf{Multimodal RewardBench 2 (MMRB2)}, the first comprehensive benchmark for reward models on multimodal understanding and (interleaved) generation. MMRB2 spans four tasks: \textbf{text-to-image, image editing, interleaved generation, and multimodal reasoning (“thinking-with-images”),} providing 1,000 expert-annotated preference pairs per task from 23 models and agents across 21 source tasks. MMRB2 is designed with: (1) practical but challenging prompts; (2) responses from state-of-the-art models and agents; and (3) preference pairs with strong human-expert consensus, curated via an ensemble filtering strategy.
Using MMRB2, we study existing judges for each subtask, including multimodal LLM-as-a-judge and models trained with human preferences. The latest Gemini 3 Pro attains 75-80\% accuracy. GPT-5 and Gemini 2.5 Pro reach 66–75\% accuracy, compared to >90\% for humans, yet surpass the widely used GPT-4o (59\%). The best performing open-source model Qwen3-VL-32B achieves similar accuracies as Gemini 2.5 Flash (64\%). 
We also show that MMRB2 performance strongly correlates with downstream task success using Best-of-N sampling and conduct an in-depth analysis that shows key areas to improve the reward models going forward. 
}

\date{\today}
\correspondence{Yushi Hu at \email{yushihu@meta.com}, Reyhane Askari-Hemmat at \email{reyhaneaskari@meta.com}, Marjan Ghazvininejad at \email{ghazvini@meta.com}}

\metadata[Code and data]{\url{https://github.com/facebookresearch/MMRB2}}
\begin{document}

\maketitle

\section{Introduction}
\label{sec:intro}

\begin{figure}[t]
    \centering
    \includegraphics[width=\linewidth]{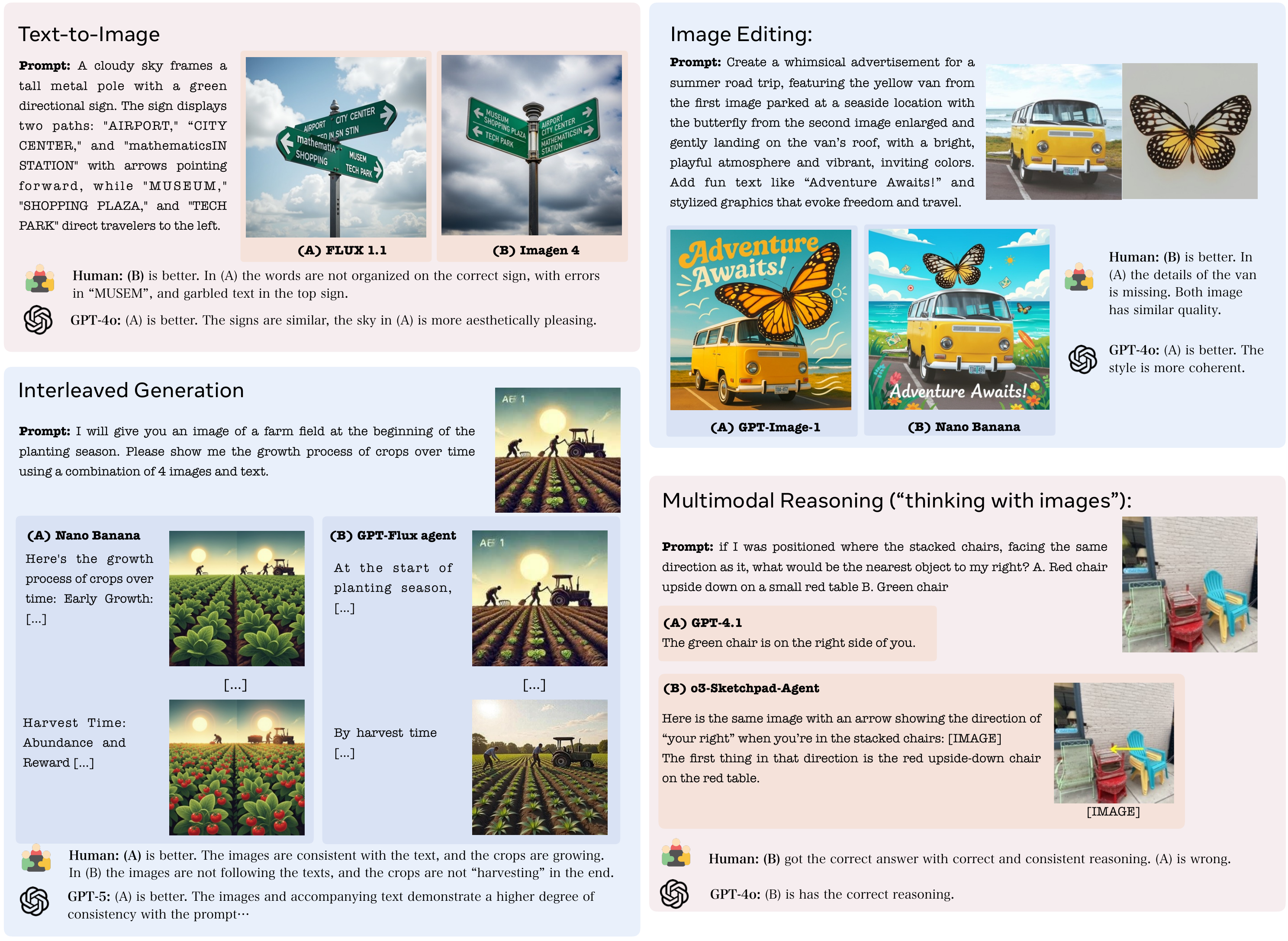}
    \caption{Examples of multimodal preference pairs in MMRB2 across four subtasks: text-to-image generation, interleaved generation, image editing, and multimodal reasoning, showing human and model judgments on challenging prompts.
    %\melissahall{"visual reasoning" to "multimodal reasoning" in the figure}
    }
    \label{fig:examples}
\end{figure}

Reward models are central to the development of LLMs~\citep{christiano2017deep, bai2022training,o1, guo2025deepseek, tulu3, yuan2024selfrewarding}. They enable scalable evaluation that tracks model performance and surfaces systematic weaknesses~\citep{zheng2023judging}. They can be used to assess data quality, which is crucial for building synthetic data pipelines~\citep{wang2022selfinstruct}.  And, as reinforcement learning becomes increasingly important in post training, high quality reward models are crucial for  surfacing or suppressing a range of different model capabilities~\citep{christiano2017deep, wu2023fine, guo2025deepseek}. Recent work has focused on developing new classes of omni models, which enable understanding, generation, and reasoning with interleaved text and images~\citep{gpt4o, chameleon2024mixed, ge2025seedxmultimodalmodelsunified, zhou2024transfusion, bagel, chen2025januspro, wang2024emu3, blip3o, Gemini25_Flash_Image}. However, reward modeling for omni models remains largely unexplored.

This omission is at least in part because there is no existing benchmark for omni reward models, making it nearly impossible to measure model quality. Unlike text-only models, omni models can generate and understand any number of texts and images together in a single arbitrarily ordered sequence. This generality creates unique challenges for reward modeling. Unlike domains such as math or coding, images are difficult to verify automatically~\citep{hessel2021clipscore, hu2023tifa, lin2024evaluating}, and high-quality preference data requires carefully designed annotation protocols~\citep{richhf}. Omni models can also be used for a very broad range of real-world applications, demanding diverse task coverage for both training and evaluation~\citep{liu2024holisticevaluationinterleavedtextandimage, ISG, yao2025mmmgcomprehensivereliableevaluation}. Finally, gathering high-quality responses needed to train and evaluate omni reward models can be challenging, since omni model capabilities are not always as strong as the models used to develop previous text-only reward benches.   

%Existing multimodal models remain further from human-level performance on their native multimodal benchmarks than state-of-the-art text-only LLMs are on text benchmarks. This gap makes it harder to reliably obtain high-quality responses needed to train and evaluate reward models.

% Existing multimodal models are also less capable than leading LLMs, \emily{this claim is not clear to me, what does less capable mean if you're measuring on different tasks? maybe it means something like, multi-modal LLMs are further from human-level performance on multi-modal tasks than LLMs are on text tasks?} making it harder to obtain “good responses” for reward model studies \MG{not 100\% clear}.
% 
% Prior work \emily{prior work on multi-modal reward model benchmarking?} has focused primarily on text-to-image generation, with a few efforts on single-image editing. To our knowledge, there is no systematic study of reward models for interleaved text–image generation or for challenging multimodal reasoning tasks that require “thinking with images.”
% 
We introduce Multimodal RewardBench 2 (MMRB2) which overcomes all of these challenges to establish a foundation for future research on omni reward modeling. 
%\MG{Should we talk about the connection with MMRB1?}\emily{+1, I think it will be confusing if it's not mentioned before MMRB2 is presented} 
MMRB2 follows Multimodal RewardBench (MMRB1)~\citep{yasunaga2025multimodal}, which covered image-text-to-text tasks for multimodal large language models (MLLMs).
MMRB2 instead covers the much more challenging case of omni models over four subtasks (Figure~\ref{fig:examples}): text-to-image, image editing, interleaved generation, and multimodal reasoning (“thinking with images”~\citep{o3}). Each subtask contains 1,000 expert-annotated preference pairs, consisting of a task prompt, a preferred response, and a rejected response. To ensure that MMRB2 is comprehensive, reliable, and highly predictive of reward model quality, we design it with three key characteristics: (1) diverse, practical, yet challenging prompts near the capability boundary of frontier models, drawn from 21 existing and newly created tasks; (2) responses generated by state-of-the-art multimodal models, ranging from SD3.5~\citep{stabilityai2024stablediffusion3_5} to GPT-Image~\citep{gpt_image} and Gemini 2.5 Flash Image~\citep{Gemini25_Flash_Image}, along with specialized agents~\citep{hu2024visualsketchpad} for interleaved generation and visual reasoning tasks where even the best models often fail; and (3) preference pairs that have >90\% agreement among human experts but which remain challenging for current judges (both MLLM-based judges and trained reward models), curated via an ensemble filtering strategy. A summary of all the prompts and multimodal models covered in MMRB2 are in Table~\ref{tab:model_prompt_source}.
%\MG{Upto here we didn't talk about filtering strategy. Maybe we need to mention 1 sentence in intro about it}.
% We build upon our earlier work, Multimodal RewardBench (MMRB1)~\citep{yasunaga2025multimodal}, which provided the first systematic evaluation framework for multimodal reward models (RMs) by benchmarking their ability to align with human preferences across vision and text. While MMRB1 offered a valuable initial step, it was limited in scope as it focuses primarily on text–image alignment and lacks coverage of interleaved and reasoning-heavy multimodal interactions.
% To address these challenges and establish a stronger foundation for future research on omni-model reward modeling, we introduce Multimodal RewardBench 2 (MMRB2). MMRB2 comprises four subtasks; text-to-image, image editing, interleaved generation, and visual reasoning (“thinking with images”). Each subtask contains 1,000 expert-annotated preference pairs, consisting of a task prompt, a preferred response, and a rejected response.

\begin{table*}[h!]
\renewcommand{\arraystretch}{1.5}  % Increased from 1.2 to 1.5 for wider row spacing
\setlength{\tabcolsep}{4pt}
\small
\centering
\begin{adjustbox}{max width=\textwidth}
\begin{tabular}{L{4cm} L{5cm} p{2cm} L{6cm} L{5cm}}
\toprule
\textbf{Category} & \textbf{Source} & \textbf{Response} & \textbf{Models} & \textbf{Task Description} \\
\midrule
Text-to-Image &
WISE~\citep{niu2025wise}\newline
EvalMuse~\citep{han2024evalmuse40kreliablefinegrainedbenchmark}\newline
OneIG-Bench~\citep{chang2025oneig}\newline
R2IBench~\citep{R2IBENCH}\newline
RealUnify~\citep{shi2025realunifyunifiedmodelstruly}
& Image &
Gemini 2.0 and 2.5 Flash Image~\citep{google2025gemini2_0_flash_image, Gemini25_Flash_Image}\newline
Imagen 3~\citep{baldridge2024imagen3}\newline
Imagen 4 and Ultra~\citep{google2025imagen4}\newline
FLUX~\citep{labs2025flux1kontextflowmatching}\newline
GPT-image-1~\citep{gpt_image}\newline
SD 3.5-L~\citep{stabilityai2024stablediffusion3_5}
& Image generation from text assessing creativity, composition, reasoning, text rendering, etc. \\
Image Editing &
DreamBench~\citep{peng2024dreambench}\newline
Emu-Edit~\citep{sheynin2024emuEDIT}\newline
HQ-Edit~\citep{hui2024hqedithighqualitydatasetinstructionbased}\newline
RISE-Bench~\citep{risebench}\newline
Text-heavy edit\newline
Multi-Image edit
& Image &
Gemini-2.0 and 2.5 Flash Image~\citep{google2025gemini2_0_flash_image, Gemini25_Flash_Image}\newline
Imagen3-Edit~\citep{baldridge2024imagen3}\newline
FLUX-Kontext~\citep{labs2025flux1kontextflowmatching}\newline
GPT-image-1~\citep{gpt_image}
& Object replacement, scene modification, style change, entity-preserving editing, reasoning-heavy editing, text-heavy editing, multi-image editing, etc. \\
Interleaved Generation &
Chameleon~\citep{chameleon2024mixed}\newline
Interleaved-Eval~\citep{liu2024holisticevaluationinterleavedtextandimage}\newline
ISG-Bench~\citep{ISG}\newline
MMMG~\citep{yao2025mmmgcomprehensivereliableevaluation}
& Text+Image &
Gemini 2.0 and 2.5 Flash Image~\citep{google2025gemini2_0_flash_image, Gemini25_Flash_Image}\newline
\textbf{Agents:}\newline
GPT-Gemini-agent\newline
GPT-GPT-image-agent\newline
GPT-Imagen-agent\newline
GPT-FLUX-agent
%~\citep{gpt41, Gemini25_Flash_Image, google2025imagen4, labs2025flux1kontextflowmatching}
& Interleaved text-image generation, storytelling, open-ended visual question answering, scene composition, 3D prediction, temporal prediction, etc. \\
Reasoning &
BLINK~\citep{fu2024blink}\newline
MindCube~\citep{MINDCUBE}\newline
VisuLogic~\citep{xu2025visulogic}\newline
V$^*$~\citep{vstar}\newline
MuirBench~\citep{wang2024muirbench}\newline
RealUnify~\citep{shi2025realunifyunifiedmodelstruly}
& Text(+Image) &
GPT-5~\citep{gpt5}\newline
GPT-4.1~\citep{gpt41}\newline
GPT-4o~\citep{gpt4o}\newline
o3~\citep{o3}\newline
Gemini 2.5 Flash~\citep{Gemini25}\newline
Gemini 2.5 Pro~\citep{Gemini25}\newline
\textbf{Sketchpad Agents~\citep{hu2024visualsketchpad}:}\newline
o3-sketchpad-agent\newline
GPT-5-sketchpad-agent
& Thinking with images, spatial reasoning, multi-image reasoning, perception-heavy tasks, etc. \\
\bottomrule
\end{tabular}
\end{adjustbox}
\caption{Overview of the four subtask categories in MMRB2, including their prompt sources, response modalities, model that were used to synthesize the data, and task descriptions. The benchmark draws from a diverse set of public and newly created datasets to cover text-to-image generation, image editing, interleaved text–image generation, and multimodal reasoning ("thinking with images").}
\label{tab:model_prompt_source}
\end{table*}

Using MMRB2, we conduct a comprehensive study of reward models for multimodal understanding and generation, including multimodal LLM-as-a-judge, task-specific metrics, and reward models trained with human preferences. Experiments show that:
\begin{itemize}
    \item MMRB2 poses significant challenges to current MLLM-as-a-judge approaches: the latest Gemini 3 Pro~\citep{gemini3pro} reaches 74-80\% accuracy across all subtasks.  GPT-5~\citep{gpt5} and Gemini 2.5 Pro~\citep{Gemini25} achieve only moderate performance (66-75\% accuracy across all subtasks) compared to >90\% for humans. The best open-source model, Qwen3-VL-32B~\citep{qwen3vl}, achieves 55\%-69\% accuracy. Notably, GPT-4o~\citep{gpt4o}, which is commonly used as an evaluator in existing benchmarks, attains only 51-65\% accuracy, suggesting that it is no longer suitable for evaluating frontier multimodal models, especially on reasoning-heavy tasks. 
    \item We study task-specific metrics (e.g., VQAScore~\citep{lin2024evaluating}) and reward models trained on human preferences (e.g., ImageReward~\citep{xu2023imagereward}, UnifiedReward~\citep{wang2025unifiedreward}), and find that they are no longer reliable on the challenging prompts and frontier models in MMRB2. For instance, VQAScore (with Qwen2.5-VL backbone) and ImageReward achieve 58.3\% and 54.0\% on text-to-image evaluation, respectively, well below MLLM-as-a-judge baselines such as Qwen3-VL-32B (64.1\%) and Gemini 3 Pro (74.4\%). While human preference training improves performance over heuristic metrics, these models still fall short of frontier MLLMs.
    %, reflecting limited multimodal generalization.
    % \emily{Maybe add some numberss here?} Although training with human preferences substantially improves judge accuracy, these trained reward models still fall far short of the best MLLM-as-a-judge results, likely because they are less generalizable than leading MLLMs.
    % 
    \item We show that performance on MMRB2 strongly correlates with performance on GenAI-Bench~\citep{GENAI-Bench}, GEdit-Bench~\citep{liu2025step1x}, ISGBench~\citep{ISG}, and EMMA~\citep{EMMA} when using different reward models for best-of-$N$ selection, suggesting that MMRB2 is a good proxy for downstream effectiveness.
    \item Further analysis of test-time scaling and fine-grained error patterns reveals substantial remaining headroom for omni model reward modeling and highlights concrete failure modes that future methods should address. Judges show notably higher agreement with human preferences on different-model pairs than on same-model pairs, with differences of up to 12\%. Moreover, in multimodal reasoning tasks, judges exhibit a strong bias toward responses that include images, with performance gaps of 27.7–49.3\% between pairs where annotators preferred image-containing responses and those where the preferred response contained only text.
\end{itemize}
Overall, MMRB2 establishes a challenging and informative benchmark that we hope will serve as a foundation for future research on omni model reward modeling, evaluation, and post-training.

% Bullet points:

% There is not even benchmarks for that \melissahall{Maybe we can call out specific tasks that are missing in existing benchmarks, for example, evaluating interleaved generations or visual reasoning.}.

% We make MMRB2, the first benchmark targeting reward models for omni model training. It contains 4 sub-tasks: text-to-image, image-editing, interleaved, and visual reasoning (with possibly thinking with image)

% Our benchmarks characteristics.

% \begin{enumerate}
%     \item Practical and challenging tasks prompts that are on the capability boundary of frontier models
%     \item Responses from SOTA models. We also build agents for the tasks that even SOTA models perform poorly.
%     \item Expert annotations. Filtering strategy to find the pairs that are consensus among human but challenging for current judges.
%     \item Comprehensive analysis on how current judges perform, how mmrb2 score correlate with downstream tasks, and their space for improvement. \melissahall{We can have a TODO to add notable findings here from our analysis, e.g. there is a gap in generation vs evaluation}
% \end{enumerate}
% \TODO{comment something about us releasing the benchmark eventually.}

\section{Related Work}
\label{sec:related_works}
\vspace{-0.05in}

\noindent\textbf{Reward modeling for visual generation.}
Building on RLHF, reward modeling has been extended beyond text. ImageReward \citep{xu2023imagereward}, HPSv2 and v3 \citep{wu2023human, ma2025hpsv3}, PickScore~\citep{kirstain2023pick} learn human preferences for text-to-image generation, improving correlation with human judgments and guiding diffusion models beyond CLIP-based proxies. For image editing, EditScore \citep{luo2025editscore} and EditReward \citep{wu2025editreward} adopt similar preference-learning frameworks. Unified approaches aim for cross-task generalization: \citet{wang2025unifiedreward} train a single multimodal reward across image, video, and understanding tasks. Despite progress, most multimodal RMs remain task-specific and lack a unified, stress-testing evaluation.

\noindent\textbf{Evaluating reward models.}
Benchmarking reward models has become an active research direction. In the text domain, RewardBench and RewardBench~2 \citep{lambert2025rewardbench,malik2025rewardbench} systematically compare LLM reward functions across diverse axes (e.g., instruction following, reasoning, safety). VL-RewardBench \citep{li2025vl} and Multimodal RewardBench \citep{yasunaga2025multimodal} assess reward models for multimodal LLM.
Llava-Critic series~\citep{xiong2025llava, wang2025llava} focus on developing reward models for these reward benchmarks. EvalPlanner \citep{saha2025evalplanner} and J1 \citep{whitehouse2025j1}  further improve reward modeling by incentivizing test-time scaling in LLM-as-a-judge.
However, existing benchmarks and judge training efforts still largely focus on image-text-to-text tasks. For image generation, researchers develop automatic evaluation metrics for text-to-image generation. CLIPScore \citep{hessel2021clipscore} offers a reference-free image--text similarity measure that correlates with human judgments but often misses compositional errors; TIFA \citep{hu2023tifa}, DSG~\citep{JaeminCho2024dsg}, and VQAScore \citep{lin2024evaluating} address this by probing alignment via VQA, improving robustness on compositional cases. OmniVerifier~\citep{zhang2025generative} further investigate on training better visual-outcome verification mdiirhtdkltrnlldrnvhbccugtflukgivodels. The human annotations collected in these works are often used as reward model evaluations.
Most existing reward model evaluations focus either on text or text-to-image generation, offering little insight into interleaved text and image.
To bridge this gap, Multimodal RewardBench 2 (MMRB2) provides a unified and challenging framework for assess reward modeling for omni models.

\section{\longmmrb}
\label{sec:benchmark}

\begin{figure}
    \centering
    \includegraphics[width=1\linewidth]{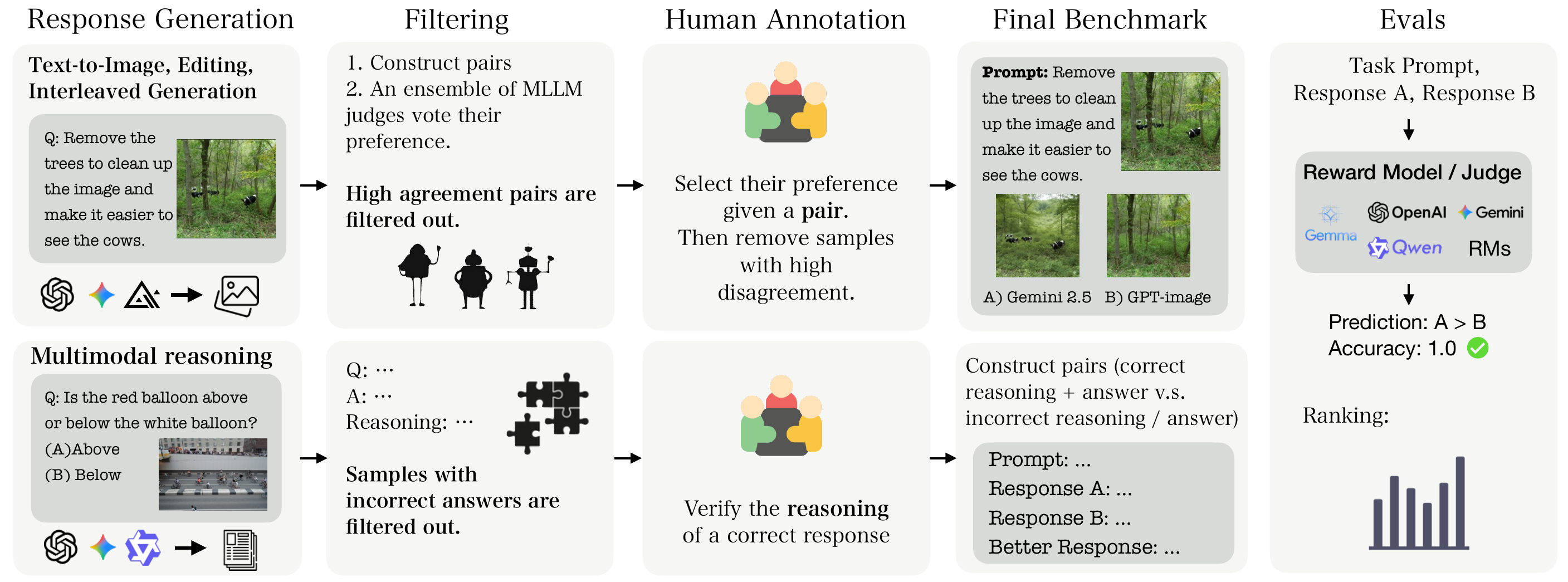}
    \caption{Overview of the MMRB2 data pipeline. The process combines ensemble MLLM judging, human verification, and multi-stage filtering to ensure high-quality, reasoning-consistent preference pairs across tasks.}
    \label{fig:annotation_pipeline}

\end{figure}

MMRB2 (Figure~\ref{fig:annotation_pipeline}) is a comprehensive omni reward model evaluation benchmark spanning a range of tasks (\S\ref{sec:benchmark:tasks}) of four types: text-to-image generation, image editing, interleaved generation, and multimodal reasoning. 
Each datapoint in MMRB2 contains a task prompt (\S\ref{sec:benchmark:prompt}) and two model responses, chosen and rejected (\S\ref{sec:benchmark:annotation}).
Reward models are evaluated based on their agreement with human annotators (\S\ref{sec:benchmark:evaluation}).

%In the following subsections, w
%We define each of the task in MMRB2 in \S\ref{sec:benchmark:tasks}. 
%We then describe the details of prompt and response collection for each task in \S\ref{sec:benchmark:prompt}, our human annotation and pair construction process in \S\ref{sec:benchmark:annotation}, and our evaluation metrics for reward models in \S\ref{sec:benchmark:evaluation}.

\subsection{Tasks in \shortmmrb}
\label{sec:benchmark:tasks}

% In \shortmmrb, each task defines a multimodal setting from which we collect candidate outputs from generators or agents. Reward models are evaluated on whether they reproduce human preferences over these candidates, given the same prompts and inputs.

\begin{figure}
    \centering
    \includegraphics[width=\linewidth]{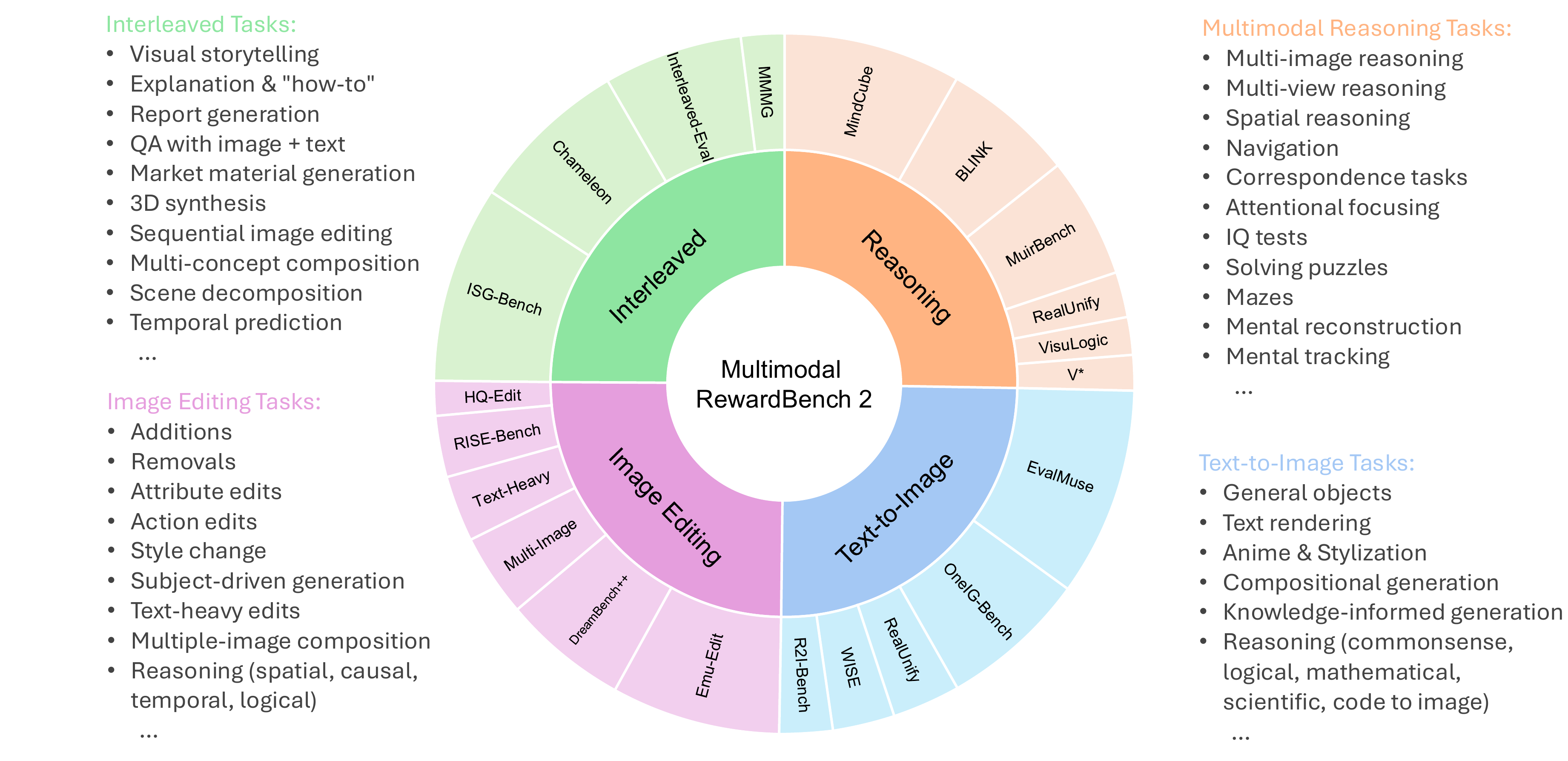}
    \caption{Breakdown of MMRB2 by task type and source, and detailed categories under each task.}
    \label{fig:piechart}

\end{figure}

\noindent\textbf{Task 1. Text-to-Image.} 
Text-to-image generation provides natural language prompts for which generators produce candidate images. 
Reward models see the prompt and the candidate images, and must prefer the human-preferred image based on factors such as object composition, spatial relationships, attribute binding, text rendering, and adherence to complex multi-object instructions.

\noindent\textbf{Task 2. Image Editing.} 
Image editing provides 1-3 input images and a textual edit instruction, along with candidate edited images from generators. 
Reward models must select the edit that best matches human preference, balancing faithfulness to the edit request with preservation of irrelevant regions. 
The edits include both single-image operations (e.g., changing attributes, scene modifications, adding/removing elements) and multi-image compositions where multiple inputs must be integrated.

\noindent\textbf{Task 3. Interleaved Generation.} 
Interleaved generation provides multimodal prompts that elicit mixed image–text sequences from generators (e.g., for storytelling, how-to guides, educational content, or multi-step reasoning). 
Reward models are asked to rank candidate interleaved outputs, capturing human preferences for coherence, global planning, and effective coordination between visual and textual content.

\noindent\textbf{Task 4. Multimodal Reasoning (Thinking with images).}
Multimodal reasoning provides complex problems that require visual understanding, logical inference, and multi-step problem solving. 
Generators may produce both text and intermediate thinking or sketchpad images; reward models must judge which candidate reasoning trajectory and final answer better aligns with human preference, emphasizing accurate perception, spatial reasoning, and clear explanation. 

See Figure~\ref{fig:examples} for examples of multimodal preference pairs in \shortmmrb{} across these four subtasks.

\subsection{Prompt and response collection}
\label{sec:benchmark:prompt}
 % For each task, we curate prompts from established benchmarks using stratified sampling to ensure diversity across difficulty levels and task categories. We only use the test split to avoid training data leaks. We also curate new tasks (for example, multi-image editing) that are practical but not covered in current benchmark.
 % We weight benchmarks by their coverage and difficulty distribution, selecting 1000 prompts per task. 
 % For each prompt, we generate multiple independent responses from 7-11 state-of-the-art models, including both API-based models and open-source alternatives.
 % We find that even the latest models like Gemini 2.5 Flash-Image often struggle on interleaved generation and multimodal reasoning. For these cases, we also build agents that can run python codes and use image generation/editing models as tools  ~\citep{hu2024visualsketchpad}.
 % Details of prompt sources and candidate models are shown in Table~\ref{tab:model_prompt_source}. Appendix~\ref{app:prompts} provides more details.
For each task, we sample prompts from existing benchmarks via stratified sampling over difficulty and subtask type, using only test splits to avoid train--test leakage. We additionally design new, practical tasks (e.g., multi-image editing) that are not covered in prior benchmarks. Benchmarks are weighted by coverage and difficulty, yielding 1{,}000 prompts per task. For each prompt, we generate multiple responses from 7--11 state-of-the-art models, including both API and open-source systems. We observe that even strong models such as Gemini 2.5 Flash Image struggle on interleaved generation and multimodal reasoning; for such cases, we further construct agents that can call Python and image generation/editing tools~\citep{hu2024visualsketchpad}. Table~\ref{tab:model_prompt_source} summarizes prompt sources and candidate models, with additional details in Appendix~\ref{app:prompts}.

\subsection{Human preference annotations}
\label{sec:benchmark:annotation}
Given prompts and responses, we developed methods to gather human preferences for each task type.

\subsubsection{Image generation, editing \& interleaved tasks} %\MG{Can we name it without numbers? :D }
% For Tasks 1-3, w
We adopt a unified annotation protocol to ensure consistency across text-to-image generation, image editing, and interleaved generation tasks. 

\noindent\textbf{Ensemble filtering.}
To focus human annotation on the most informative comparisons, we first apply an ensemble filtering pipeline that removes easy preference pairs where one response is almost unanimously preferred.
We collect judgments from nine multimodal judges spanning API models (GPT-5, GPT-4.1, GPT-4o, Gemini 2.5 Flash, Gemini 2.5 Pro) and open-source VL models (Gemma-3-27B/12B/4B, Qwen-2.5-VL-7B), covering a wide range of capability.
Each judge evaluates every pair twice, once in forward order (A vs.\ B) and once in reverse order (B vs.\ A), to mitigate position bias (see Appendix~\ref{app:prompts} for the exact prompts).

We define easy pairs as those where the majority label appears in at least 90\% of all judge evaluations across both orderings, and discard them because they provide little signal about fine-grained differences between reward models.
While ensemble filtering can in principle introduce bias, the diversity of the judges and the high 90\% threshold restrict filtering to near-trivial cases and mitigate systematic bias from any single model.

\noindent\textbf{Human Preference Annotation.} We employed professional annotators via the Surge AI platform to collect high-quality human preferences.\footnote{Annotators were compensated at an hourly rate of \$85.} Each pair is independently evaluated by three annotators who have no knowledge of which model generated each response. Annotators assess each response using a comprehensive evaluation framework with different criteria tailored to each task category. Finally for each pair, annotators provide their overall preference for answer A vs B on a 7-point Likert scale where we convert these ratings to preferences. See details in Appendix~\ref{app:prompts}. 
%using the following mapping: ratings 5-7 indicate preference for A, ratings 1-2 indicate preference for B, and ratings 3-4 are treated as ties. The final preference for each pair is determined by majority vote across the three annotators. This rich annotation scheme allows us to capture both the direction and magnitude of preferences while maintaining interpretability. \MG{Can we provide exact 7-point scale? It is hard to follow. }
 We implement several additional quality control measures. First, we filter out annotations with high inter-annotator disagreement, specifically removing pairs where the rating spread (maximum rating minus minimum rating) exceeds 4 points on the 7-point scale. We also exclude ambiguous annotations where the average rating falls too close to the scale midpoint (within the 3.0-4.0 range), as these indicate genuine uncertainty rather than clear preferences from human annotations. Finally, we remove pairs where the majority vote results in a tie, as these provide limited signal for evaluating judge agreement. 
 
For the three generative tasks, we collected approximately 17,700 human preference judgments, each evaluated by three independent annotators, resulting in 5900 judgments overall. After filtering, we retain 1,000 pairs per task (approximately 50\% of the initial set). Inter-annotator agreement on these filtered pairs is high: 95.6\% overall (excluding ties), with task-specific rates of 95.3\% for image generation, 96.3\% for image editing, and 95.2\% for interleaved generation. %Ties are relatively frequent (14.5\% of annotator pairs), reflecting the subtle distinctions near the frontier of current model capabilities. This filtering ensures that MMRB2’s generative tasks provide reliable and challenging signal for evaluating reward models and MLLM judges, while focusing human annotation effort on the most informative comparisons.

% eight questions. Each category has its own questions.
% Get good annotations

% In the end, we get
% \begin{itemize}
%     \item Task 1. Text-to-Image. What questions, agreement
%     \item Task 2.
%     \item Task 3.
% \end{itemize}

\subsubsection{Multimodal reasoning task} 
% \MG{Again let's name it. I think it needs rewriting, not 100\% clear why this task is different? filtering? what did we ask annotators to do (Maybe ref the exact questions we are asking annotators in appendix?) ? and then how did we make the pairs?  how we combine 3 annotations? (The details are less than task 1-3)}
Because multimodal reasoning prompts have ground truth answers, we collect human annotations per model response (rather than pairwise) then construct pairs.

\noindent\textbf{Human annotation.} 
We filter generated model responses from \S\ref{sec:benchmark:prompt} to those that contain both the correct answer to the prompt and some form of reasoning.
We then balance samples across responses that include text-only reasoning and those that reason with both images and text. 
We collect three human annotations per response that indicate whether the reasoning contained in the model response is correct.
The annotator instructions are listed in Appendix~\ref{sup:task_breakdown}.

\noindent\textbf{Pair construction.}
With the annotated responses, we construct preference pairs.
For the human-preferred sample of each pair, we select model responses in which all three human annotators agree that the reasoning contains no major errors and the model answer is correct. 
For the non-preferred sample of each pair, we utilize two kinds of responses: \textit{Correct answer, incorrect reasoning}, where the model answer is correct but all three annotators consider the reasoning to contain major errors, and \textit{Incorrect answer, with reasoning}, where the model answer is incorrect and some form of reasoning is included. 
For each pair, the two model responses may share the same modality (both text-only or both image+text) or be a combination. 
No model response is duplicated across pairs. 
For more details, see Appendix ~\ref{sup:task_breakdown}.
% The full breakdown of samples between the pair types and modalities can be found in Appendix ~\ref{sup:task_breakdown}.
% \MG{Maybe we can provide some statistics here about final percentages}

\subsection{Evaluation Method}
\label{sec:benchmark:evaluation}

Finally, we use the preference pairs to evaluate reward models on MMRB2.

\noindent\textbf{Positional consistent dual evaluation} 
Position bias is common problem; models have a systematic preference for the first item in a pair~\citep{min-etal-2022-noisy, tan2025judgebenchbenchmarkevaluatingllmbased}. To mitigate this, each pair is evaluated twice per judge: once in its original order (A vs. B) and once with responses swapped (B vs. A). Both forward and reverse judgments are retained as independent data points, doubling judge-human comparison instances. This protocol improves agreement statistics by increasing sample size and penalizes judges with high position bias.
% , since inconsistent judgments across orderings naturally reduce alignment with human preferences.
% \MG{Mention that this is a known problem and maybe cite different papers, I can cite} For pairwise judges, to mitigate any position bias the judge may have, each pair is evaluated twice per judge: first in its original ordering (Response A vs. Response B), then with responses swapped (Response B vs. Response A). This dual evaluation protocol allows us to detect and quantify position bias (the tendency of a judge to systematically favor the first or second position regardless of content quality). When processing reverse-order evaluations, we automatically swap the predicted preferences to align with the original ordering (i.e., if the judge selects "A" in the reverse evaluation where B is presented first, we record this as preferring "B" in the original ordering).
% For each pair, a judge produces two independent judgments: one forward and one reverse. Both judgments are retained and treated as separate evaluation instances when computing agreement with human annotations, effectively doubling the number of judge-human comparison points. This approach provides two benefits: (1) it yields more robust agreement statistics by increasing sample size, and (2) it naturally penalizes judges with high position bias, as inconsistent forward/reverse judgments will disagree with human preferences at different rates.

\noindent\textbf{Judge-Human Agreement Computation} We measure judge-human agreement by comparing each judge evaluation against the human preference for the corresponding pair. Human preference is determined by majority vote across three annotators for Tasks 1-3 and unanimous agreement of reasoning and answer correctness in Task 4.
% \MG{we are still using some kind of majority voting, but much stronger version}. 
For each judge evaluation (whether forward or reverse), we compute a binary agreement score: 1.0 if the judge's preference matches the human preference (including tie-to-tie matches), and 0.0 otherwise. %Crucially, we do not assign partial credit for tie prediction, a judge prediction of "tie" only receives credit when humans also selected tie as the majority preference. 

% The overall agreement rate is then computed by 
% averaging these pair-level scores across all pairs, which is equivalent to:

% $$\frac{\sum_{\text{pairs}} \sum_{\text{evals}} \mathbb{1}[\text{judge}_{\text{eval}} = \text{human}_{\text{majority}}]}{\text{Total Evaluations}} \times 100\%$$

% where the numerator sums agreement across all forward and reverse evaluations, and the denominator is the total number of evaluations (typically $2 \times N$ for $N$ pairs). This produces a percentage representing how often the judge agrees with human judgment.

\section{Experiments}
\label{sec:experiments}

We conduct a comprehensive study of omni reward modeling with MMRB2 along a number of dimensions: evaluation of MLLM-as-a-judge (\S\ref{sec:experiments:mllm}), evaluation of other task-specific evaluators (\S\ref{sec:experiments:other_evaluators}), and in-depth analysis on various aspects of the benchmark and omni model reward modeling (\S\ref{sec:experiments:correlation} - \ref{sec:experiments:reward_model_ensemble}).

\subsection{Performance of MLLM-as-a-judge} 
\label{sec:experiments:mllm}

\begin{table}[h]
\centering
\resizebox{0.8\linewidth}{!}{%
\begin{tabular}{lccccc}

\toprule[1.2pt]

Judge & 
\begin{tabular}[c]{@{}c@{}} Text to\\ Image\end{tabular} & 
\begin{tabular}[c]{@{}c@{}} Image\\ Editing\end{tabular} & 
\begin{tabular}[c]{@{}c@{}} Interleaved\\ Generation\end{tabular} &
\begin{tabular}[c]{@{}c@{}} Multimodal\\ Reasoning\end{tabular} &
Avg. \\

\midrule[1.2pt]

\multicolumn{6}{c}{Open-source multimodal LLMs} \\

\midrule

Gemma 3 4B~\citep{gemma}           & 51.7 & 51.0 & 51.3 & 48.8 & 50.7 \\
Gemma 3 12B~\citep{gemma}          & 56.0 & 58.0 & 58.0 & 49.3 & 55.3 \\
Gemma 3 27B~\citep{gemma}          & 58.3 & 60.2 & 61.1 & 49.4 & 57.3 \\
Qwen2.5-VL-7B~\citep{qwen25vl}     & 50.4 & 57.1 & 48.4 & 47.5 & 50.9 \\
Qwen2.5-VL-72B~\citep{qwen25vl}    & 59.1 & 64.6 & 62.3 & 50.0 & 59.0 \\
Qwen3-VL-8B~\citep{qwen3vl}        & 59.4 & 61.7 & 61.5 & 54.6 & 59.3 \\
Qwen3-VL-32B~\citep{qwen3vl}       & 64.1 & 67.3 & 70.5 & 56.6 & 64.6 \\
Qwen3-VL-30BA3B~\citep{qwen3vl}    & 60.0 & 59.5 & 57.3 & 57.3 & 58.5 \\
Qwen3-VL-235BA22B~\citep{qwen3vl}  & 62.0 & 64.8 & 69.0 & 55.9 & 62.9 \\

\midrule

\multicolumn{6}{c}{API-based Models} \\

\midrule

GPT-4o~\citep{gpt4o}               & 60.3 & 65.0 & 61.5 & 51.9 & 59.7 \\
GPT-4.1~\citep{gpt41}              & 65.8 & 68.2 & 67.0 & 53.0 & 63.5 \\
GPT-5~\citep{gpt5}                 & \underline{70.5} & \underline{73.8} & 74.4 & \underline{70.2} & \underline{72.2} \\
Gemini 2.5 Flash~\citep{Gemini25}    & 63.1 & 66.5 & 69.4 & 57.5 & 64.1 \\
Gemini 2.5 Pro~\citep{Gemini25}      & \underline{70.5} & 71.3 & \underline{75.1} & 66.6 & 70.9 \\
Gemini 3 Pro~\citep{gemini3pro}    & \textbf{74.4} & \textbf{74.9} & \textbf{76.4} & \textbf{79.5} & \textbf{76.3} \\

\bottomrule[1.2pt]
\end{tabular}
}
\caption{MLLM-as-a-judge accuracies on MMRB2. The best numbers are \textbf{bolded} and the second best are \underline{underlined}. Gemini 3 Pro is the best across all tasks. Qwen3-VL-32B is the best open-source model.}
\label{tab:mllm_results_main}
\end{table}

\noindent\textbf{Setup.} We evaluate all tasks on API-based models GPT-4o, GPT-4.1, GPT-5, Gemini 2.5 Flash, Gemini 2.5 Pro, Gemini 3 Pro and open-source models Qwen 2.5-VL (7B and 72B)~\citep{qwen25vl}, Qwen 3-VL (8B, 32B, 30BA3B,  235BA22B)~\citep{qwen3vl} and Gemma 3 (4B, 12B, and 27B)~\citep{gemma}. For each task type, we design task-specific evaluation prompts with detailed rubrics (see Appendix~\ref{app:prompts}). We follow the positional consistent dual evaluation method in \S\ref{sec:benchmark:evaluation} to mitigate positional bias.

\noindent\textbf{Results.} Table~\ref{tab:mllm_results_main} reveals substantial variation in judge-human agreement across models and tasks. API-based models generally outperform open-source alternatives, with \textbf{Gemini 3 Pro achieving the strongest overall performance across all tasks.} 
%GPT-5 demonstrates exceptional capabilities on image editing (73.8\%) and multimodal reasoning (70.2\%), while Gemini 2.5 Pro achieves the highest agreement on interleaved generation (75.1\%). Both models are tied for the best performance on image generation (70.5\%). 
GPT-5 and Gemini 2.5 Pro also achieves decent accuracy on text-to-image generation, image editing, and interleaved generation (70 - 75\% accuracy). 
Notably, multimodal reasoning proves to be the most challenging task across all models except Gemini 3 Pro, with even top API models achieving only 52-70\% agreement on reasoning tasks (compared to 63-75\% on multimodal generation tasks). 
This difficulty may stem from multiple valid solution paths, varying levels of explanation detail that humans may value differently, or the challenge of assessing both correctness and reasoning quality simultaneously.

Gemma 3, Qwen2.5-VL, and Qwen3-VL families of models all \textbf{perform better on MMRB2 as number of parameters scales.}
Additionally, the performance gap between API-based and open-source models has narrowed with recent open-source advances. 
The top API models (Gemini 3 Pro, Gemini 2.5 Pro, GPT-5) achieve agreement rates of 65-80\% across most tasks, while the best open-source models now reach competitive performance levels.
Qwen3-VL-32B is the strongest open-source model, achieving 64.1-70.5\% across tasks.
Notably, its 70.5\% agreement rate for interleaved generation approaches API-based model performance. 
While the Qwen3-VL series generally outperforms the Gemma 3 and Qwen2.5 families on image-related tasks, even some of the Gemma 3 and Qwen2.5 variants are within a few percentage points of API-based models. 
However, open-source models still show large gaps with API-based models on multimodal reasoning: the strongest, Qwen3-VL 30BA3B at 57.3\%, trails Gemini 3 Pro by 22 percentage points.%, while other open-source models have even lower performance (47-49\%).
% This suggests that strong reasoning capabilities may be prerequisite for reliably judging reasoning quality in others.

% Reasoning evaluation proves most difficult across all judges, with substantially lower agreement rates.  \MG{it is repetetive. We have "Notably, multimodal reasoning proves to be ...}.

% Different models exhibit varying patterns of task-specific strengths. GPT-5 excels at interleaved and editing tasks (73-75\%) but shows relatively weaker performance on image generation (68\%). GPT-4o demonstrates high variance across tasks (51-64\%), suggesting less robust generalization. In contrast, Gemini 2.5 Pro maintains stable performance across all modalities, making it potentially more reliable as a general-purpose evaluator. This consistency may be valuable for benchmark creation where uniform evaluation quality across diverse content types is essential. 

% \MG{This is also a little bit repetitive. We mentioned GPT-5 demonstrates exceptional capabilities on image ... before}

% \MG{we can mention 50\% accuracy is random guessing. }

\subsection{Performance of supervised reward models}
\label{sec:experiments:other_evaluators}

\begin{table}[h]
    \centering
\resizebox{0.8\linewidth}{!}{%    
\begin{tabular}{lccc}
\toprule[1.2pt]
Judge & Text to Image & Image Editing* & Multimodal Reasoning*\\
% \begin{tabular}[c]{@{}c@{}}Text to\\Image\end{tabular} & 
% \begin{tabular}[c]{@{}c@{}}Image\\Editing*\end{tabular} & 
% \begin{tabular}[c]{@{}c@{}}Multimodal\\Reasoning*\end{tabular} \\
\midrule[1.2pt]
\multicolumn{4}{c}{MLLM-as-a-judge}\\
\midrule
Qwen2.5-VL-7B~\citep{qwen25vl} & 50.4 & 57.8 & 53.7 \\
Qwen3-VL-32B~\citep{qwen3vl} & 64.1 & 66.4 & 69.9 \\
GPT-5~\citep{gpt5}       & 70.5 & 74.3 & 83.8 \\
\midrule
\multicolumn{4}{c}{CLIP-based evaluators} \\
\midrule
CLIPScore~\citep{hessel2021clipscore}   & 51.0 & - & - \\
ImageReward~\citep{xu2023imagereward}    & 54.0 & - & - \\
HPSv2~\citep{hpsv2}          & 54.7 & - & - \\
PickScore~\citep{kirstain2023pick}      & 58.6 & - & - \\
\midrule
\multicolumn{4}{c}{Qwen2.5-VL-7B-based evaluators} \\
\midrule
VQAScore~\citep{lin2024evaluating}    & 58.3 & - & - \\
HPSv3~\citep{ma2025hpsv3}          & 60.2 & - & - \\
EditReward~\citep{wu2025editreward}    & - & 67.2 & - \\
% LlavaCritic~\citep{xiong2025llava}    & - & - & 48.5 \\
UnifiedReward~\citep{wang2025unifiedreward}  & 59.8 & - & 55.1  \\
\bottomrule[1.2pt]

\end{tabular}
}
\caption{Other evaluators' accuracies on MMRB2. Note that all task-specific evaluators except CLIPScore and VQAScore have been trained with human preference pairs. *For editing we use the single-image subset; for reasoning we use the text-only-output subset, ensuring fair comparison among evaluators. }
\label{tab:mllm_results_task}
\end{table}

Besides directly prompting MLLMs to act as judges, prior work has proposed a range of automatic metrics and preference-trained reward models targeting the tasks in MMRB2. We evaluate these methods on the three MMRB2 tasks—\emph{text-to-image} generation, \emph{image editing}, and \emph{multimodal reasoning}. To the best of our knowledge, there are currently no evaluators specifically tailored for interleaved text–image outputs.

\noindent\textbf{Setup.}
Unless otherwise noted, we adopt the default metaprompt provided by each official library.
For text-to-image, we consider two families of evaluators. The first is CLIP-based~\citep{radford2021learning}, including CLIPScore~\citep{hessel2021clipscore} and its preference-trained variants ImageReward~\citep{xu2023imagereward}, HPSv2~\citep{hpsv2}, and PickScore~\citep{kirstain2023pick}.  
The second family is based on Qwen2.5-VL-7B~\citep{qwen25vl}. We evaluate VQAScore~\citep{lin2024evaluating}, which scores generated images using model logits, as well as the preference-trained reward models HPSv3~\citep{ma2025hpsv3} and UnifiedReward~\citep{wang2025unifiedreward}. We evaluate all of the above models on the MMRB2 text-to-image task.
Qwen2.5-VL-7B has also been used as the backbone for reward models on other tasks, including EditReward~\citep{wu2025editreward} for image editing and UnifiedReward~\citep{wang2025unifiedreward} for multimodal understanding. Because EditReward is trained only on single-image editing, and UnifiedReward is trained on single-image image-to-text tasks, we evaluate them on the corresponding single-image subsets of MMRB2 to ensure a fair comparison among evaluators.
% we evaluate the widely used  and VQAScore~\citep{lin2024evaluating} (Qwen2.5-VL-7B variant), as well as preference-trained models including  HPSv3~\citep{ma2025hpsv3}, , and UnifiedReward~\citep{wang2025unifiedreward} (Qwen-based).
% For \textbf{image editing}, we evaluate the latest EditReward~\citep{wu2025editreward} (Qwen-based) on the single-image editing subset of MMRB2, which matches its training scope.
% For \textbf{multimodal reasoning}, existing reward models are predominantly trained on single-image queries with text-only outputs; we therefore evaluate LlavaCritic~\citep{xiong2025llava} and UnifiedReward~\citep{wang2025unifiedreward} on the corresponding MMRB2 subset. 
% For LlavaCritic, we tested two metaprompts and report the stronger result; full prompt details are provided in \S\ref{app:prompts}. 
% \noindent\textbf{Results}
Table~\ref{tab:mllm_results_task} summarizes the results. 
%Some entries are directly using pre-trained models, while others are models trained with human preference data. 
% We also include representative MLLM-as-a-judge baselines for context. 
%\MG{ What do you mean by directly using pre-trained model? are we talking about MLLM as a judge?}

\noindent\textbf{Preference training substantially improves reward-model accuracy.}
Several reward models share the same base architecture as our MLLM baselines (e.g., EditReward, UnifiedReward, and HPSv3 are based on Qwen2.5-VL-7B), and some are CLIP-based (ImageReward, HPSv2, PickScore). Relative to the Qwen2.5-VL-7B judge, EditReward yields a \textbf{+9.4\%
} gain on editing (57.8~$\rightarrow$~67.2), and UnifiedReward improves text-to-image by \textbf{+9.4\%} (50.4~$\rightarrow$~59.8) and reasoning by \textbf{+1.4\%} (53.7~$\rightarrow$~55.1). 
Similarly, compared to CLIPScore (51.0), CLIP-based preference models show consistent gains: ImageReward 54.0 (\textbf{+3.0}~\%), HPSv2 54.7 (\textbf{+3.7}~\%), and PickScore 58.6 (\textbf{+7.6}~\%).
These results indicate that training with human preferences is an effective way to boost evaluator performance on multimodal tasks.
%\MG{Can we present the results in the table better, so it is easer to see the base model for each? }

\noindent\textbf{Reward models can be out-of-distribution; strong MLLMs remain strong judges.}
% \noindent\textbf{Strong MLLMs remain strong judges}
Despite the above gains, most preference-trained reward models still underperform a larger open-source judge such as Qwen3-VL-32B across tasks; a notable exception is EditReward, which is competitive on editing (67.2 vs.\ 66.4). One plausible explanation is a distribution shift: several reward models were trained on data from earlier-generation systems (e.g., SD~2.1–era), and their accuracy diminishes when judging outputs from more capable, recent models. 
Overall, newer reward models (HPSv3, EditReward, UnifiedReward) are far better than older ones, yet stronger MLLM still set a high bar through simple prompting. 
% This suggests that (i) continued refreshment of preference data is important for keeping reward models in-distribution, and (ii) choosing a strong base model can yield larger gains than preference training alone.

% \noindent\textbf{Multimodal reasoning}
% For visual reasoning, we evaluate LLaVA-Critic~\citep{xiong2025llava} and UnifiedReward~\citep{wang2025unified}.
% We find that the default LLaVA-Critic meta-prompt often considers both model responses of equal quality, while an alternative meta-prompt that encourages the judge to select one response improves performance. 
% We report results with the higher performing meta-prompt in the main portion of the paper and include both meta-prompts and their results in Appendix \TODO{}.
% \TODO{Not much better than random, do worse than MLLM as judges}

\subsection{Correlation with downstream tasks}
\label{sec:experiments:correlation}

\begin{figure}[h]
    \centering
    \includegraphics[width=\linewidth]{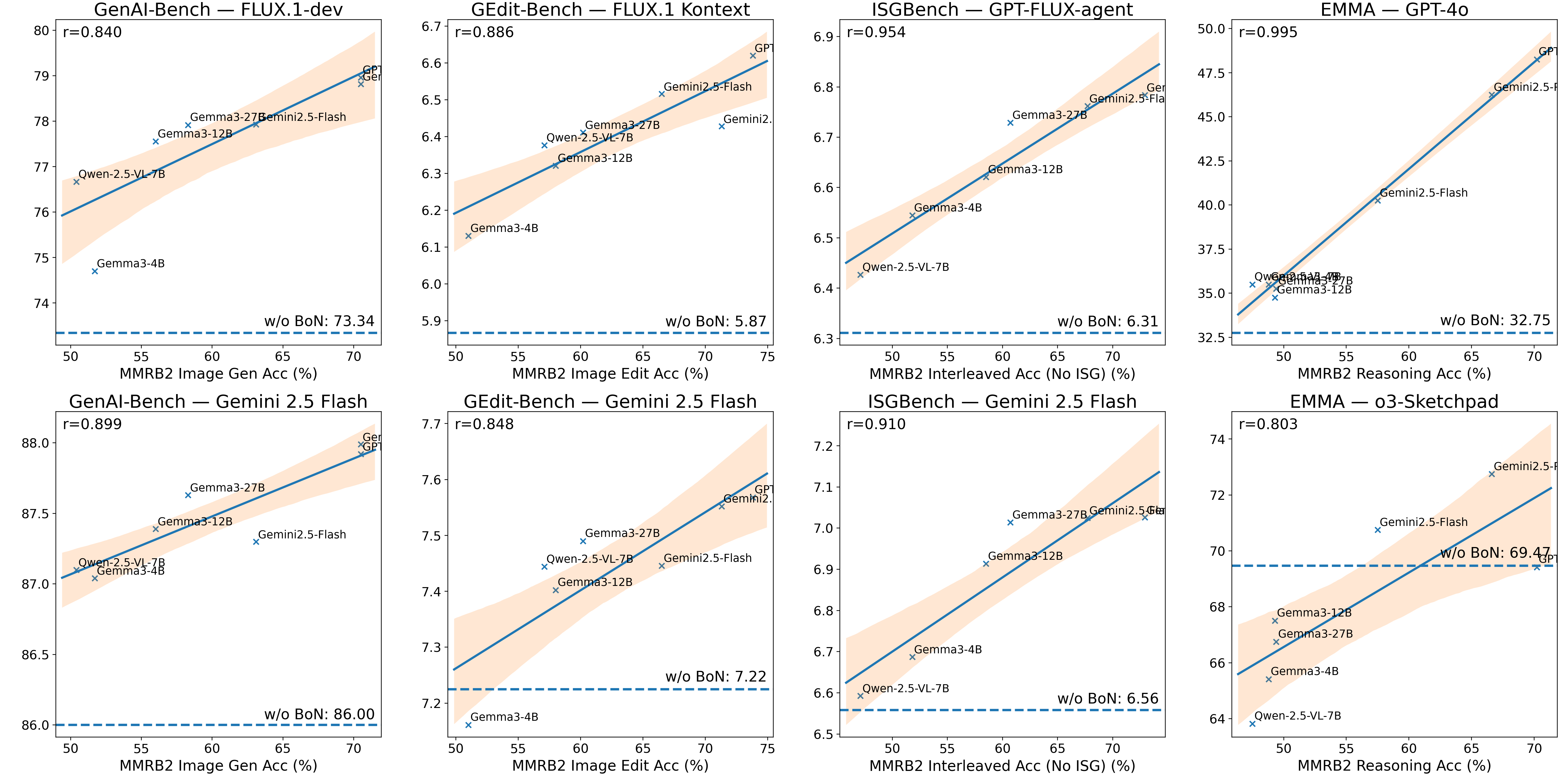}
    \caption{Downstream best-of-N score v.s. MMRB2 performance. We perform best-of-N sampling with 2 base models each on 4 tasks (GenAI-Bench~\citep{GENAI-Bench}, GEdit-Bench~\citep{liu2025step1x}, ISG-Bench~\citep{ISG}, and EMMA~\citep{EMMA}). A judge's score on MMRB2 strongly correlates with improvement in downstream tasks when it is used in best-of-N sampling, highlighting MMRB2's utility for downstream task success.}
    \label{fig:best_of_n}

\end{figure}

A key research question is whether MMRB2 performance can predict downstream task performance. To address this question, following prior works~\citep{lightman2023let, li2025vl}, we conduct best-of-N sampling with different rewards. We experiment with 4 tasks: GenAI-Bench~\citep{GENAI-Bench}, GEdit-Bench~\citep{liu2025step1x}, ISG-Bench~\citep{ISG}, and EMMA~\citep{EMMA}, each corresponds to one task in MMRB2. For each query, we generate $N=8$ candidate responses from two base models, one strong model and a weaker one, and we use 7 different MLLM-as-a-judge to select the best one via knockout tournaments. Then we evaluate the selected response with each task's metrics. 

Figure~\ref{fig:best_of_n} shows all the results. The x-axis is the MMRB2 performance, and Y-axis is the score of the best-of-N response selected by different rewards. We exclude GPT-4o and 4.1 because they are often used as evaluators. For interleaved generation, we remove ISGBench preference pairs when computing MMRB2 scores to avoid leakage. The results show that there is a strong correlation between best-of-N performance and MMRB2 scores. A good reward model can give great gains on downstream performance, even with the simple best-of-N sampling. For example, FLUX's GenAI-Bench score improved from 73\% to 79\%, and GPT-4o's accuracy on EMMA improved from 32\% to 45\%,  when using GPT-5 as best-of-N selector. We can still see consistent gains even for strong base models like Gemini 2.5 Flash Image and o3-Sketchpad. The strong correlation (>0.8 Pearson's $r$ for all tasks and models) between MMRB2 and downstream task performance validates that MMRB2 is a good proxy for downstream effectiveness.

\subsection{Fine-grained analysis of errors}
% \label{sec:experiments:fine_grained_analysis}

% \begin{figure}
%     \centering
%     \includegraphics[width=\linewidth]{figs/sample_error_analysis.png}
%     \caption{A placeholder for error analysis. borrowed from VLRB}
%     \label{fig:error_analysis}
% \end{figure}

\noindent\textbf{Same-model pairs \textit{vs.} different-model pairs.}
Our benchmark contains 57.4\% same-model pairs (comparing two outputs from the same model) and 42.6\% different-model pairs (comparing outputs from different models), allowing us to assess judge performance across both scenarios. See results in Table~\ref{tab:judge_detailed_main}.

\begin{table}[h]
\centering
\resizebox{0.7\linewidth}{!}{%
% \begin{tabular}{llccc}
% \toprule
% \textbf{Task} & \textbf{Judge} & \textbf{Overall (\%)} & \textbf{Same-M (\%)} & \textbf{Dif-M (\%)} \\
% \midrule
% \multirow{5}{*}{Image Gen.} & Gemini 2.5 Pro & 71.4 & 67.8 & 74.7 \\
% & GPT 5 & 68.0 & 63.7 & 71.9 \\
% & GPT 4.1 & 67.0 & 60.3 & 73.1 \\
% & Gemini 2.5 Flash & 64.3 & 59.2 & 68.8 \\
% & GPT 4o & 63.1 & 59.9 & 66.0 \\
% \midrule
% \multirow{5}{*}{Image Editing} & GPT 5 & 73.8 & 70.6 & 76.4 \\
% & Gemini 2.5 Pro & 71.1 & 65.0 & 76.1 \\
% & GPT 4.1 & 67.6 & 65.6 & 69.3 \\
% & Gemini 2.5 Flash & 66.8 & 62.6 & 70.2 \\
% & GPT 4o & 64.0 & 60.3 & 67.1 \\
% \midrule
% \multirow{5}{*}{Interleaved} & GPT 5 & 75.2 & 70.5 & 79.9 \\
% & Gemini 2.5 Pro & 73.3 & 69.5 & 77.0 \\
% & Gemini 2.5 Flash & 70.5 & 66.6 & 74.3 \\
% & GPT 4.1 & 65.8 & 61.9 & 69.7 \\
% & Gemma 3 27B & 61.1 & 59.9 & 62.2 \\
% \bottomrule
% \end{tabular}
\begin{tabular}{llccc}
\toprule
\textbf{Task} & \textbf{Judge} & \textbf{Overall (\%)} & \textbf{Same-M (\%)} & \textbf{Diff-M (\%)} \\
\midrule
\multirow{5}{*}{Image Generation} 
& Gemini 3 Pro & 74.4  &70.4 & 79.7 \\
& Gemini 2.5 Pro & 70.5 & 68.4 & 73.4 \\
& GPT-5 & 70.5 & 66.8 & 75.6 \\
& GPT-4.1 & 65.8 & 61.6 & 71.4 \\
& Qwen3-VL-32B & 64.1 & 59.1 & 71.0 \\
% & Gemini 2.5 Flash & 63.1 & 59.9 & 67.4 \\
\midrule
\multirow{5}{*}{Image Editing} 
& Gemini 3 Pro &  74.9&    71.0&     79.3\\
& GPT-5 & 73.8 & 71.7 & 76.2 \\
& Gemini 2.5 Pro & 71.3 & 66.7 & 76.6 \\
& GPT-4.1 & 68.2 & 65.6 & 71.3 \\
& Qwen3-VL-32B & 67.3 & 64.5 & 70.5 \\
% & Gemini 2.5 Flash & 66.5 & 62.5 & 71.2 \\
\midrule
\multirow{5}{*}{Interleaved} 
& Gemini 3 Pro & 76.4&   72.8&     82.0 \\
& Gemini 2.5 Pro & 75.1 & 70.7 & 81.9 \\
& GPT-5 & 74.4 & 69.4 & 82.1 \\
& Qwen3-VL-32B & 70.5 & 66.7 & 76.4 \\ 
& Gemini 2.5 Flash & 69.4 & 65.0 & 76.3 \\
% & Qwen3 VL 235BA22B & 69.0 & 67.8 & 70.8 \\
\midrule
\multirow{5}{*}{Reasoning} 
& Gemini 3 Pro & 79.5&     78.7&     79.8\\
& GPT-5 & 70.2 & 68.4 & 70.8 \\
& Gemini 2.5 Pro & 66.6 & 70.5 & 65.4 \\
& Gemini 2.5 Flash & 57.5 & 59.9 & 56.7\\
& Qwen3-VL-30BA3B & 57.3 & 57.4 & 57.3 \\
% & Qwen3 VL 32B & 56.6 & 59.6 & 55.7 \\
\bottomrule
\end{tabular}
}
\caption{Detailed performance breakdown of top 5 judges per task showing overall agreement, same-model pairs, and different-model pairs with human preferences.}
\label{tab:judge_detailed_main}
\end{table}

Across the image generation, editing, and interleaving tasks, we observe a consistent pattern for all judges: judges achieve higher agreement with human on different-model pairs compared to same-model pairs. 
For the best-performing judges, this gap ranges from 5-13 percentage points. 
For example, on image generation, Gemini 3 Pro achieves 79.7\% agreement on different-model pairs but only 70.4\% on same-model pairs (9.3 point gap).
% , while Gemini 2.5 Pro shows 73.4\% vs. 68.4\% (5.0 point gap). 
% The pattern holds across all tasks: for image editing, GPT-5 scores 76.2\% on different-model pairs versus 71.7\% on same-model pairs (4.5 point gap), while Gemini 2.5 Pro shows a larger 9.9 point gap (76.6\% vs. 66.7\%). 
% For interleaved generation, the gaps are even more pronounced, GPT-5 achieves 82.1\% versus 69.4\% (12.7 point gap) and Gemini 2.5 Pro shows 81.9\% versus 70.7\% (11.2 point gap). 
This pattern holds across tasks: same-model pairs demand fine-grained judgments within one model’s outputs, while different-model pairs reveal larger gaps rooted in capability differences.
% The consistent 5-13 point performance gap across all judges (including frontier models like GPT-5) suggests current reward models struggle with subtle discrimination. 
% This highlights that effective reward models must excel at both coarse-grained comparison (easy) and fine-grained within-model assessment (hard), which remains an open challenge.

\noindent\textbf{Same-modality pairs \textit{vs} mixed-modality pairs.}
For the multimodal reasoning task, we study how judges perform differentially for pairs constructed with responses from the same modality (e.g., text response \textit{vs.} text response) versus mixed modalities (e.g., text response \textit{vs.} text-image response). 
Full results are reported in Table~\ref{tab:supp_8_modalities}.

\begin{table}[h]
\resizebox{\linewidth}{!}{%
\begin{tabular}{lccllcccc} \toprule[1.2pt] 
\textbf{}                  & \multicolumn{2}{c}{\textbf{\begin{tabular}[c]{@{}c@{}}Same modality: \\ Image+text\end{tabular}}}   & \multicolumn{2}{c}{\textbf{\begin{tabular}[c]{@{}c@{}}Same modality:\\ Text\end{tabular}}}                                                  & \multicolumn{2}{c}{\textbf{\begin{tabular}[c]{@{}c@{}}Mixed modality:\\ Pref: Image+text; Not Pref: Text\end{tabular}}}                                                           & \multicolumn{2}{c}{\textbf{\begin{tabular}[c]{@{}c@{}}Mixed modality:\\ Pref: Text; Not Pref: Image+text\end{tabular}}}                                                           \\ \midrule
\textbf{Model}                  & \textbf{\begin{tabular}[c]{@{}c@{}}Correct  vs.\\  incorrect reason\end{tabular}} & \textbf{\begin{tabular}[c]{@{}c@{}}Correct vs.\\  incorrect answer\end{tabular}} & \multicolumn{1}{c}{\textbf{\begin{tabular}[c]{@{}c@{}}Correct vs.\\  incorrect reason\end{tabular}}} & \multicolumn{1}{c}{\textbf{\begin{tabular}[c]{@{}c@{}}Correct  vs.\\  incorrect answer\end{tabular}}} & \textbf{\begin{tabular}[c]{@{}c@{}}Correct  vs.\\  incorrect reason\end{tabular}} & \textbf{\begin{tabular}[c]{@{}c@{}}Correct  vs.\\  incorrect answer\end{tabular}} & \textbf{\begin{tabular}[c]{@{}c@{}}Correct  vs.\\  incorrect reason\end{tabular}} & \textbf{\begin{tabular}[c]{@{}c@{}}Correct  vs.\\  incorrect answer\end{tabular}} \\ \midrule
\multicolumn{9}{c}{Open-source models} \\
\midrule
% \textbf{Model}             & \textbf{} & \textbf{} & \multicolumn{1}{c}{\textbf{}} & \multicolumn{1}{c}{\textbf{}} & \textbf{} & \textbf{} & \textbf{} & \textbf{} \\
% balanced\_llama32-11b      & 49.12     & 45.97     & \multicolumn{1}{c}{45.09}     & \multicolumn{1}{c}{47.25}     & 44.32     & 46.94     & 41.35     & 39.00     \\
% balanced\_llama4-scout-17b & 44.35     & 40.97     & \multicolumn{1}{c}{49.55}     & \multicolumn{1}{c}{53.99}     & 51.14     & 40.00     & 30.19     & 34.50     \\
% balanced\_qwen25vl72b      & 50.88     & 48.73     & \multicolumn{1}{c}{53.12}     & \multicolumn{1}{c}{55.25}     & 78.41     & 69.00     & 16.04     & 23.47     \\
Gemma3 4B                  & 47.39     & 50.63     &   \multicolumn{1}{c}{50.00} &    \multicolumn{1}{c}{48.32} & 63.64     & 57.00     & 38.68     & 36.00     \\
Gemma3 12B                 & 49.57     & 51.47     &  \multicolumn{1}{c}{54.02} &  \multicolumn{1}{c}{52.10} & 81.82     & 73.50     & 15.09     & 11.50     \\
Gemma3 27B                 & 51.30     & 50.21     &  \multicolumn{1}{c}{51.79} &  \multicolumn{1}{c}{51.68} & 87.50     & 79.50     & 10.38     & 10.50     \\
Qwen2.5-VL-7B                 & 49.12     & 48.10     &  \multicolumn{1}{c}{51.34} & \multicolumn{1}{c}{50.00} & 52.27     & 39.00     & 47.17     & 40.31     \\
Qwen2.5-VL-72B                & 52.63     & 48.10     & \multicolumn{1}{c}{53.57}  &  \multicolumn{1}{c}{54.41}  & 78.41     & 68.00     & 16.04     & 23.98     \\
Qwen3-VL-8B        & 57.46     & 52.53     & \multicolumn{1}{c}{58.48}     & \multicolumn{1}{c}{54.20}     & 71.59     & 73.00     & 34.91     & 36.73     \\
Qwen3-VL-32B       & 62.28     & 54.43     & \multicolumn{1}{c}{60.71}     & \multicolumn{1}{c}{56.93}     & 78.41     & 80.00     & 25.47     & 33.16     \\
Qwen3-VL-30BA3B    & 58.77     & 55.72     & \multicolumn{1}{c}{57.59}     & \multicolumn{1}{c}{56.30}     & 75.00     & 78.00     & 36.79     & 43.37     \\
Qwen3-VL-235BA22B  & 58.11     & 57.02     & \multicolumn{1}{c}{55.80}     & \multicolumn{1}{c}{57.14}     & 85.23     & 81.96     & 23.58     & 26.02     \\
\midrule
\multicolumn{9}{c}{API-based models} \\
\midrule
% unifiedreward-und          & NaN       & NaN       &                               &                               & NaN       & NaN       & NaN       & NaN       
GPT-4o                      & 50.43     & 50.42     &    \multicolumn{1}{c}{55.80} &    \multicolumn{1}{c}{56.51}  & 81.82     & 80.00     & 18.87     & 18.00     \\
GPT-4.1                      & 56.09     & 50.42     &  \multicolumn{1}{c}{58.04} &  \multicolumn{1}{c}{58.61}  & 93.18     & 81.50     & 10.38     & 13.50     \\
GPT-5                       & 69.57     & 67.02     &   \multicolumn{1}{c}{75.89}  &  \multicolumn{1}{c}{80.25} & 88.64     & 88.00     & 36.79     & 40.00     \\ 
Gemini 2.5 Flash             & 60.53     & 58.47     & \multicolumn{1}{c}{56.25}  & \multicolumn{1}{c}{59.03} & 86.36     & 76.00     & 16.98     & 38.42     \\
Gemini 2.5 Pro                & 73.91     & 66.18     & \multicolumn{1}{c}{62.95}  & \multicolumn{1}{c}{65.55}   & 84.09     & 79.00     & 43.40     & 58.00     \\
Gemini 3 Pro                 &  71.88      &   84.75    &  \multicolumn{1}{c}{75.45}    &   \multicolumn{1}{c}{82.49}  &  \multicolumn{1}{c}{84.88} &  87.00   &  66.98   &   72.00       \\ 
% llama32-11b                & 24.32     & 34.43     &                               &                               & 20.00     & 43.75     & 21.43     & 58.33     \\
% llavacritic                & NaN       & NaN       &                               &                               & NaN       & NaN       & NaN       & NaN       \\

                                                                             \bottomrule[1.2pt]
\end{tabular}
}
\caption{Multimodal reasoning performance breakdown by pair modality and pair type.}
\label{tab:supp_8_modalities}
\end{table}

We find that for mixed-modal pairs, all judges exhibit a strong bias towards the response that contains images. 
This is true even of the highest performing models: the accuracy of GPT-5 for mixed-modal pairs when the preferred response contains an image is 49.3 points higher than pairs where the preferred response contains text (88.2\% vs. 38.9\%), and Gemini 2.5 Pro and Qwen3-VL-30BA3B have gaps of 27.7 and 36.0 points respectively.  Gemini 3 Pro performs much better, but still has a 17.9 point gap.
Additionally, we find that this trend holds for both pair types: those constructed with an incorrect response \textit{vs.} a correct response and those with incorrect reasoning \textit{vs.} correct reasoning.  %judges tend to perform similarly between these two pair types overall.
% \MG{a little bit hard to follow, can you rewrite?}
% Finally, we find that nearly all judges perform similarly between same modality pairs constructed from text-only responses and pairs constructed from text-image responses, with the exception of GPT5, which performs better for text-only pairs by 11.1 points.

\subsection{Test-time scaling of rewards}
\label{sec:experiments:reward_model_ensemble}

\begin{figure}[h]
\centering
\includegraphics[width=0.6\linewidth]{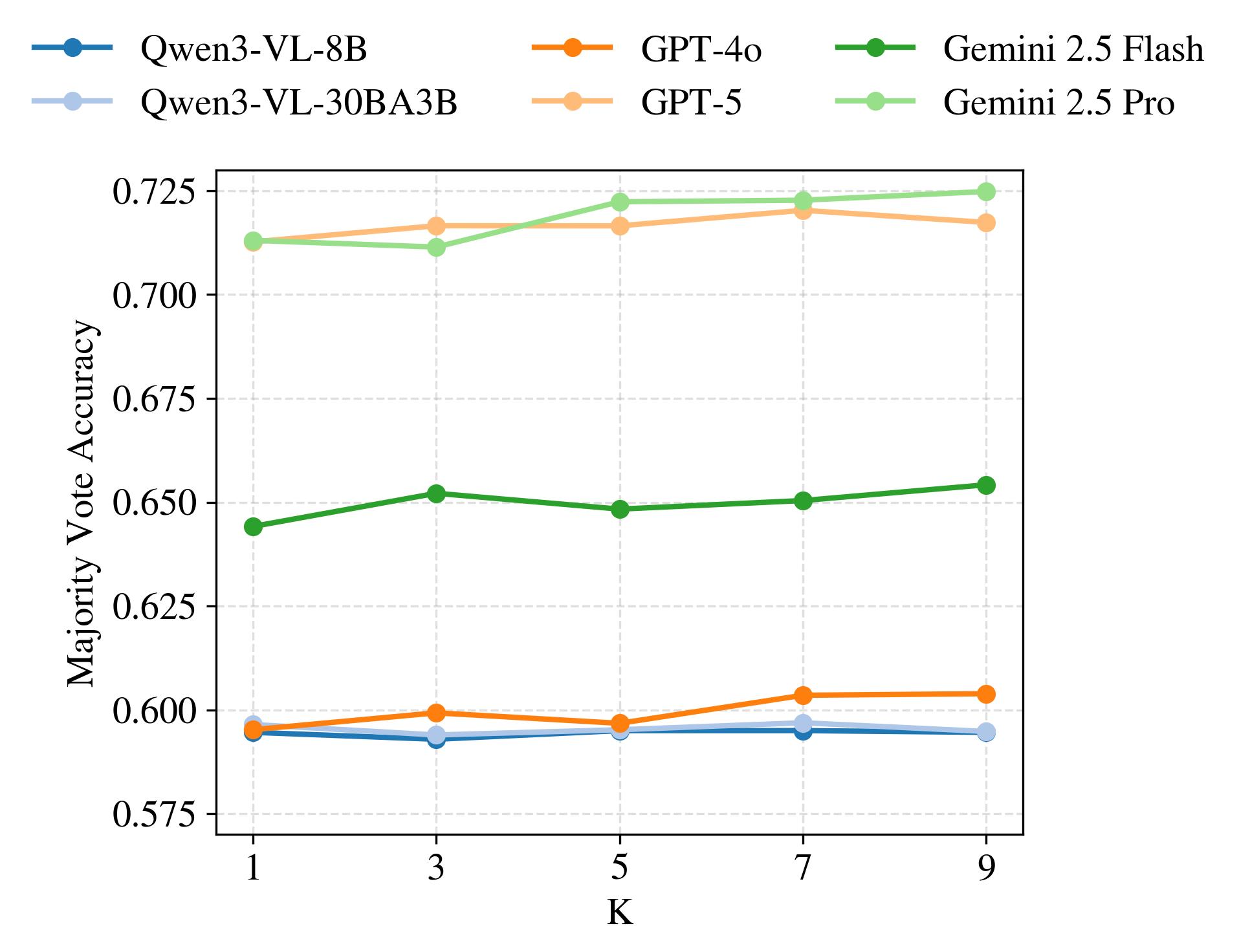}
\caption{Majority-vote accuracy of each MLLM as the number of samples $K$ varies. Test-time scaling yields small gains for GPT and Gemini models but no improvement for Qwen3-VL.}
\label{fig:test_time_scaling}

\end{figure}

Prior work~\citep{wang2022self,brown2024large} shows that test-time scaling can substantially improve LLM performance. We ask whether similar gains transfer to multimodal reward models. For each judge, we draw $K \in \{1,3,5,7,9\}$ independent judgments and take a majority vote as the final decision. We report majority-vote accuracy averaged over the four MMRB2 tasks (300 examples per task) in Fig.~\ref{fig:test_time_scaling}.

The effects are model-dependent, echoing trends in prior observations~\citep{li2025vl}. Qwen3-VL models show no measurable improvement as $K$ increases. In contrast, GPT-4o, Gemini 2.5 Flash, GPT-5, and Gemini 2.5 Pro improve by roughly $0.8$–$1.2\%$  at $K{=}9$, with Gemini 2.5 Pro showing the largest gain (from $71.3\%$ to $72.5\%$). Overall, test-time scaling provides only modest returns for multimodal reward models compared with text-only LLMs, suggesting that alternative scaling methods are needed for multimodal rewards.

\section{Conclusion}

We introduce Multimodal RewardBench 2, the first comprehensive benchmark for omni reward models spanning four tasks: text-to-image, image editing, interleaved generation, and multimodal reasoning. 
Our analysis suggests that current omni reward models, particularly the latest Gemini 3 Pro, can serve as proxies for human evaluation on multimodal generation tasks, achieving 74-80\% agreement. 
However, the substantial disagreement remaining (20-26\%) indicates that human evaluation remains essential, , and that other models, including GPT-5, lag significantly behind Gemini 3 Pro.
Overall, MMRB2 establishes a challenging and informative benchmark that we hope will serve as a foundation for future research on omni model reward modeling, evaluation, and post-training. 

\noindent\textbf{Limitations and future extensions.}
As the first comprehensive benchmark targeting omni reward models, MMRB2 focuses on core settings and overall human preferences in interleaved text--image scenarios. The construction pipeline is modular and can be extended to additional evaluation dimensions (e.g., safety- and bias-sensitive preferences), richer task formats (e.g., multilingual tasks, in-the-wild prompts, multi-turn and agentic interactions), and further modalities (e.g., video and audio). Further discussion is provided in Appendix~\ref{app:limitations}.

\section{Acknowledgements}

We would like to thank Mason Yu, Christophe Ropers, Nate Ekberg, Cynthia Gao, Justin Hovey, Jaimie Hsu, Samantha Snowden, and all annotators from Surge AI for their invaluable contributions to data annotation. We also thank Jonea Gordon and Vanessa Stark for their assistance with the approval process. Additionally, we are grateful to  Mary Williamson, Xiaochuang Han, Adriana Romero Soriano, Michal Drozdzal, Xudong Wang, Michihiro Yasunaga, Ishan Misra, Amita Kamath, Inna Lin, and Karen Chen for their insightful discussions and support throughout this project.

\clearpage
\newpage
\bibliographystyle{assets/plainnat}
\bibliography{paper}

\clearpage
\newpage
\beginappendix

\noindent\textbf{Contents of Supplementary Material}\par

\begingroup
\setlength{\parindent}{0pt}
\newcommand{\appitem}[2]{%
  \hyperref[#1]{#2}\dotfill\pageref{#1}\par}
  
\appitem{app:experiments}{Additional experimental results}
\appitem{supp:annotation}{Details of annotation and pair construction}
\appitem{app:prompts}{Additional details for prompts, response generation, and MLLM-as-a-judge}
\appitem{app:limitations}{Limitations and Future Directions}
\appitem{app:examples}{Examples}

\endgroup

\section{Additional Experimental Results}
\label{app:experiments}

\subsection{Performance by task source and pair type}

The pairwise evaluation results presented in Tables~\ref{tab:image_generation_pairwise}, \ref{tab:image_editing_pairwise},  \ref{tab:interleaved_pairwise}, \ref{tab:reasoning_pairwise} distinct performance patterns across multimodal tasks and model capabilities. 

\textbf{Image Generation}: Performance varies moderately across benchmarks, with realunify (48-78.5\%) and oneigbench (73-74\% for top models) showing higher judge agreement rates, while wise consistently yields the lowest scores across all models (47-66\%). 

\textbf{Image Editing}: The breakdown reveals stark differences in benchmark difficulty, with text-based editing benchmarks (text: 54-83\%, risebench: 48-83\%) showing significantly higher agreement rates compared to general editing tasks (emu-edit: 49-72\%, multi-image editing: 51-71\%). This pattern holds consistently across all judge models, indicating that text rendering or text-focused editing provides clearer discriminative signals for pairwise evaluation than open-ended creative edits.

\textbf{Interleaved}: Performance is relatively uniform across benchmarks for top models, with isgbench consistently scoring highest (76-79\% for frontier models) and all benchmarks clustering within a 5-8 percentage point range for leading judges. 

\textbf{Reasoning}: This task exhibits the most dramatic benchmark-level variance, with muirbench showing substantially higher scores (36-76\%) compared to other benchmarks, while vstar proves exceptionally challenging (30-52\%). The performance on blink and mindcube clusters tightly (44-55\% for most models), suggesting these represent a baseline reasoning difficulty.

\begin{table}[h]
\centering
\resizebox{0.7\linewidth}{!}{%
\begin{tabular}{lcccccc}
\toprule[1.2pt]
\textbf{Judge Model} & \textbf{Overall \%} & \textbf{evalmuse} & \textbf{oneigbench} & \textbf{r2ibench} & \textbf{realunify} & \textbf{wise} \\
 & & \textbf{(n=390)} & \textbf{(n=278)} & \textbf{(n=128)} & \textbf{(n=93)} & \textbf{(n=111)} \\
\midrule
Gemini 3 Pro           & 74.4\% & 74.5\% & 74.5\%& 72.7\% & 80.1\% & 71.2\%  \\
Gemini 2.5 Pro         & 70.5\% & 69.6\% & 73.0\% & 70.7\% & 77.4\% & 61.7\% \\
GPT5                & 70.5\% & 67.3\% & 73.2\% & 72.7\% & 78.5\% & 66.2\% \\
GPT 4.1               & 65.8\% & 62.4\% & 69.6\% & 66.8\% & 75.8\% & 58.6\% \\
Qwen-3 vl32b          & 64.1\% & 63.3\% & 68.3\% & 60.9\% & 65.6\% & 59.0\% \\
Gemini25flash       & 63.1\% & 60.3\% & 64.4\% & 62.1\% & 75.3\% & 60.8\% \\
Qwen-3 vl235ba22b     & 62.0\% & 59.2\% & 65.5\% & 60.2\% & 68.8\% & 59.0\% \\
GPT 4o               & 60.3\% & 58.1\% & 65.5\% & 59.4\% & 65.1\% & 52.3\% \\
Qwen-3 vl30ba3b       & 60.0\% & 57.6\% & 64.2\% & 57.8\% & 59.1\% & 61.3\% \\
Qwen-3 vl8b           & 59.4\% & 59.2\% & 62.8\% & 57.4\% & 62.4\% & 50.9\% \\
Qwen25vl72b         & 59.1\% & 56.8\% & 62.1\% & 56.2\% & 67.7\% & 55.9\% \\
Gemma 3-27b          & 58.3\% & 56.7\% & 59.5\% & 60.2\% & 66.7\% & 51.8\% \\
Llama4-17b    & 56.7\% & 56.3\% & 58.1\% & 53.4\% & 58.0\% & 58.1\% \\
Gemma 3-12b          & 56.0\% & 54.6\% & 58.6\% & 53.1\% & 62.9\% & 52.3\% \\
Gemma 3-4b           & 51.7\% & 50.1\% & 53.2\% & 50.8\% & 54.3\% & 52.7\% \\
Qwen25vl7b          & 50.4\% & 48.8\% & 55.9\% & 47.7\% & 48.4\% & 46.8\% \\
\bottomrule[1.2pt]
\end{tabular}
}
\caption{Image Generation: Pairwise model evaluation breakdown by benchmark.}
\label{tab:image_generation_pairwise}
\end{table}

\begin{table}[h]
\centering
\resizebox{0.8\linewidth}{!}{%
\begin{tabular}{lcccccccc}
\toprule[1.2pt]
\textbf{Judge Model} & \textbf{Overall \%} & \textbf{dreambench} & \textbf{emu-edit} & \textbf{hq-edit} & \textbf{multi-image editing} & \textbf{risebench} & \textbf{text} \\
 & & \textbf{(n=242)} & \textbf{(n=329)} & \textbf{(n=53)} & \textbf{(n=178)} & \textbf{(n=84)} & \textbf{(n=114)} \\
\midrule
Gemini 3 Pro   & 74.9\% & 70.0\% & 75.8\%  & 79.2\% & 68.5\% & 82.1\% & 84.6\%   \\
GPT5                & 73.8\% & 71.3\% & 71.6\% & 72.6\% & 71.1\% & 82.7\% & 83.3\% \\
Gemini 2.5 Pro         & 71.3\% & 70.2\% & 68.4\% & 76.0\% & 64.9\% & 79.8\% & 83.3\% \\
GPT 4.1               & 68.2\% & 67.6\% & 63.4\% & 68.9\% & 66.0\% & 74.4\% & 82.0\% \\
Qwen-3 vl32b          & 67.3\% & 63.6\% & 65.8\% & 72.6\% & 64.9\% & 71.4\% & 77.6\% \\
Gemini25flash       & 66.5\% & 63.8\% & 67.6\% & 67.9\% & 59.6\% & 72.6\% & 74.6\% \\
Qwen-3 vl235ba22b     & 66.0\% & 66.7\% & 64.1\% & 67.0\% & 57.9\% & 72.0\% & 77.6\% \\
GPT 4o               & 65.0\% & 66.5\% & 60.6\% & 60.4\% & 64.6\% & 66.1\% & 76.3\% \\
Qwen25vl72b         & 64.6\% & 63.6\% & 63.5\% & 61.3\% & 60.1\% & 63.7\% & 79.4\% \\
Qwen-3 vl8b           & 61.7\% & 59.3\% & 61.9\% & 61.3\% & 54.8\% & 62.5\% & 76.3\% \\
Llama4-17b    & 61.1\% & 60.2\% & 59.8\% & 65.3\% & 57.3\% & 64.4\% & 67.9\% \\
Gemma 3-27b          & 60.2\% & 61.2\% & 60.2\% & 58.5\% & 57.6\% & 53.0\% & 68.4\% \\
Qwen-3 vl30ba3b       & 59.5\% & 58.1\% & 58.5\% & 59.4\% & 56.5\% & 61.9\% & 68.4\% \\
Gemma 3-12b          & 58.0\% & 58.9\% & 55.0\% & 56.6\% & 57.6\% & 60.7\% & 64.0\% \\
Qwen25vl7b          & 57.1\% & 59.5\% & 53.8\% & 57.5\% & 53.7\% & 52.4\% & 70.2\% \\
Gemma 3-4b           & 51.0\% & 53.3\% & 49.1\% & 49.1\% & 51.4\% & 47.6\% & 54.4\% \\
\bottomrule[1.2pt]
\end{tabular}
}
\caption{Image Editing: Pairwise model evaluation breakdown by benchmark.}
\label{tab:image_editing_pairwise}
\end{table}

\begin{table}[h]
\centering
\resizebox{0.7\linewidth}{!}{%
\begin{tabular}{lcccccc}
\toprule[1.2pt]
\textbf{Judge Model} & \textbf{Overall \%} & \textbf{chameleon} & \textbf{interleavedeval} & \textbf{isgbench} & \textbf{mmmg} \\
 & & \textbf{(n=284)} & \textbf{(n=267)} & \textbf{(n=421)} & \textbf{(n=28)} \\
\midrule
Gemini 3 Pro         &  76.4\% & 76.4\% & 76.8\% & 76.1\% & 76.8\%  \\
Gemini 2.5 Pro        & 75.1\% & 73.4\% & 71.5\% & 78.5\% & 75.0\% \\
GPT5               & 74.4\% & 72.9\% & 72.8\% & 76.5\% & 71.4\% \\
Qwen-3 vl32b         & 70.5\% & 66.9\% & 70.4\% & 73.3\% & 66.1\% \\
Gemini25flash      & 69.4\% & 64.4\% & 70.8\% & 71.5\% & 75.0\% \\
GPT 4.1              & 67.0\% & 65.3\% & 66.3\% & 69.1\% & 60.7\% \\
Qwen-3 vl235ba22b    & 66.7\% & 63.5\% & 66.3\% & 68.9\% & 69.6\% \\
Qwen25vl72b        & 62.3\% & 59.9\% & 61.4\% & 64.7\% & 58.9\% \\
GPT 4o              & 61.5\% & 60.9\% & 60.1\% & 63.3\% & 53.6\% \\
Qwen-3 vl8b          & 61.5\% & 59.3\% & 63.7\% & 61.3\% & 66.1\% \\
Gemma 3-27b         & 61.1\% & 59.9\% & 59.4\% & 62.4\% & 69.6\% \\
Gemma 3-12b         & 58.0\% & 57.6\% & 58.1\% & 58.2\% & 58.9\% \\
Qwen-3 vl30ba3b      & 57.3\% & 55.8\% & 56.8\% & 57.7\% & 69.6\% \\
Llama4-scout-17b   & 54.4\% & 55.7\% & 54.9\% & 52.5\% & 66.7\% \\
Gemma 3-4b          & 51.3\% & 50.4\% & 52.8\% & 51.1\% & 50.0\% \\
Qwen25vl7b         & 48.4\% & 48.4\% & 46.4\% & 49.8\% & 48.2\% \\
\bottomrule[1.2pt]
\end{tabular}
}
\caption{Interleaved: Pairwise model evaluation breakdown by benchmark.}
\label{tab:interleaved_pairwise}
\end{table}

\begin{table}[h]
\centering
\resizebox{0.8\linewidth}{!}{%
\begin{tabular}{lcccccccc}
\toprule[1.2pt]
\textbf{Judge Model} & \textbf{Overall \%} & \textbf{blink} & \textbf{mindcube} & \textbf{muirbench} & \textbf{realunify} & \textbf{visulogic} & \textbf{vstar} \\
 & & \textbf{(n=355)} & \textbf{(n=367)} & \textbf{(n=137)} & \textbf{(n=55)} & \textbf{(n=49)} & \textbf{(n=37)} \\
\midrule
Qwen-3 vl32b         & 56.6\% & 52.4\% & 55.6\% & 71.5\% & 56.4\% & 61.2\% & 45.9\% \\
Qwen-3 vl30ba3b      & 56.5\% & 54.6\% & 52.9\% & 70.3\% & 56.9\% & 62.1\% & 51.1\% \\
Qwen-3 vl235ba22b    & 55.9\% & 52.6\% & 54.2\% & 76.1\% & 52.8\% & 59.2\% & 30.6\% \\
% UnifiedReward-UND    & 55.1\% & 67.4\% & -- & -- & 47.4\% & 57.9\% & 28.6\% \\
Qwen-3 vl8b          & 53.7\% & 51.9\% & 50.9\% & 64.2\% & 54.4\% & 60.4\% & 48.5\% \\
Qwen25vl72b          & 50.2\% & 46.7\% & 52.0\% & 57.7\% & 50.9\% & 54.1\% & 29.7\% \\
% Llama32-11b          & 45.6\% & 44.7\% & 47.1\% & 43.7\% & 50.9\% & 41.8\% & 43.2\% \\
Llama4-scout-17b     & 44.5\% & 43.5\% & 47.7\% & 35.8\% & 49.1\% & 44.9\% & 48.6\% \\
\bottomrule[1.2pt]
\end{tabular}
}
\caption{Reasoning: Pairwise model evaluation breakdown by benchmark. }
\label{tab:reasoning_pairwise}
\end{table}

\subsection{Win rate analysis on generations}
\label{sec:experiment:win_rates}
We also report the generation capabilities of MLLMs as content producers (models generating the multimodal content being evaluated, reported in Table~\ref{tab:model_performance}). Judging requires discriminative understanding and alignment with human preferences, while generation requires creative synthesis and technical execution. A model may excel at one role while underperforming at the other, as we observe in our results.

Table~\ref{tab:model_performance} presents the win rates of generative models across MMRB2's three generation tasks (Tasks 1--3), where win rate is computed as $(wins + 0.5 \times ties) / \text{total}$ comparisons based on human majority preferences. These are the same model outputs that judges evaluate in Table~\ref{tab:mllm_results_main}, allowing us to assess both generation quality and judgment accuracy within a unified framework.

\noindent\textbf{Image Generation.} GPT-Image-1 (60.4\%) narrowly leads text-to-image generation, closely followed by Imagen 4 (57.4\%), Imagen 4 Ultra (56.5\%), and Gemini 2.5 Flash (54.3\%), indicating a highly competitive landscape among top proprietary models with less than 6 points separating the leaders. Open-source models lag substantially: Stable Diffusion 3.5 Large (41.0\%) and FLUX (36.8\%) trail by 19--24 points.

\noindent\textbf{Image Editing.} Interestingly, general-purpose multimodal models such as Gemini 2.5 Flash (59.2\%) and GPT-Image-1 (53.2\%) outperform specialized models. While Imagen Edit achieves only a 35.2\% win rate despite being purpose-built for editing, the gap is less severe than earlier reports suggested. FLUX-Kontext (49.0\%) demonstrates competitive performance for an open-source solution, though it still trails the leaders. These results suggest that strong vision--language understanding provides significant advantages for instruction-based editing, even if specialized architectures are not entirely obsolete.

\noindent\textbf{Interleaved Generation.} Agent-based systems dominate, with GPT-Gemini Agent (57.1\%) and GPT-Image Agent (56.9\%) leading by narrow margins. Native multimodal models like Gemini 2.5 Flash (53.2\%) perform competitively, narrowing the gap with agent architectures. GPT-FLUX Agent's improved but still modest performance (40.4\%) confirms that agent quality depends critically on component model quality, though the improvement suggests that better integration strategies can help.

% \noindent\textbf{Key Insights} (1) Top proprietary models cluster tightly (54-60\% win rates), indicating mature and competitive text-to-image generation with diminishing returns from specialization. (2) The proprietary-open-source gap persists at 15-20 points for generation, with the largest gap in text-to-image (19-24 points), though this represents a narrowing trend. (3) For multi-modal applications requiring both generation and evaluation capabilities, versatile models like Gemini 2.5 Flash offer more consistent value across tasks, while specialized models show higher variance in performance.
\begin{table}[h]
\renewcommand{\arraystretch}{1.2}
\setlength{\tabcolsep}{4pt}
\small
\centering
\resizebox{0.5\columnwidth}{!}{%
\begin{tabular}{L{1cm} L{2cm} L{3cm} L{2cm}}
\toprule
\textbf{Rank} & \textbf{Task} & \textbf{Model} & \textbf{Win Rate (\%)} \\
\midrule
\multirow{8}{*}{\parbox{1cm}{1\\2\\3\\4\\5\\6\\7\\8}}
& \multirow{8}{*}{Image Gen.}
& GPT-Image-1               & 60.4 \\
& & Imagen 4                  & 57.4 \\
& & Imagen 4 Ultra            & 56.5 \\
& & Gemini 2.5 Flash          & 54.3 \\
& & Imagen 3                  & 49.2 \\
& & Gemini 2.0 Flash          & 45.6 \\
& & SD 3.5 Large & 41.0 \\
& & FLUX                      & 36.8 \\
\midrule
\multirow{5}{*}{\parbox{1cm}{1\\2\\3\\4\\5}}
& \multirow{5}{*}{Image Editing}
& Gemini 2.5 Flash          & 59.2 \\
& & GPT-Image-1               & 53.2 \\
& & FLUX-Kontext              & 49.0 \\
& & Gemini 2.0 Flash          & 47.1 \\
& & Imagen Edit               & 35.2 \\
\midrule
\multirow{6}{*}{\parbox{1cm}{1\\2\\3\\4\\5\\6}}
& \multirow{6}{*}{Interleaved}
& GPT-Gemini Agent          & 57.1 \\
& & GPT-Image Agent           & 56.9 \\
& & Gemini 2.5 Flash          & 53.2 \\
& & Gemini 2.0 Flash          & 46.2 \\
& & GPT-Imagen Agent          & 42.1 \\
& & GPT-FLUX Agent            & 40.4 \\
\bottomrule
\end{tabular}
}%
\caption{Model win rates (\%) on \longmmrb~ranked by performance within each task. Win rate is computed as (wins + 0.5 × ties) / total comparisons.}
\label{tab:model_performance}
\end{table}

\section{Details for Annotation and Pair Construction}
\label{supp:annotation}

\subsection{Tasks 1-3}

\begin{figure}
    \centering
    \includegraphics[width=0.7\linewidth]{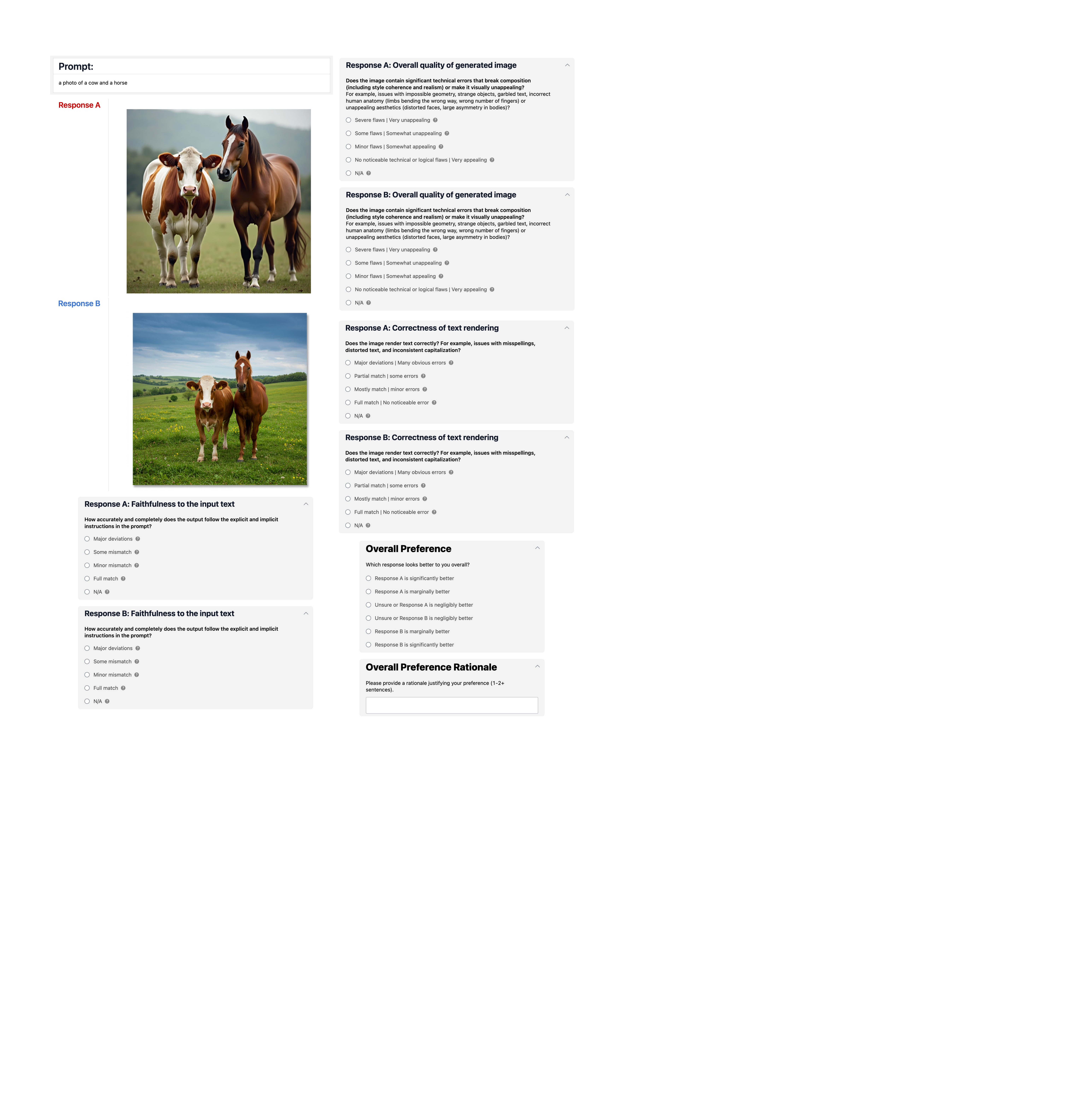}
    \caption{Annotation interface for the MMRB2 text-to-image task. Note that for image editing and interleaved tasks, there are more fine-grained questions.}
    \label{fig:annotation_fig123}
\end{figure}

Figure~\ref{fig:annotation_fig123} shows a sample of the annotation interface for the MMRB2 text-to-image task. In this section we provide additional details on the human annotation procedure.

For each annotation task, we provide a prompt and two responses, A and B, and the goal is to assess the quality of each response and then rate them. Annotators answer the following questions:

\begin{itemize}
    \item \textbf{Prompt Quality Check:} \\
    \hspace{1em}Indicate whether the prompt is correct (\textbf{Yes/No}).
    \item \textbf{Pointwise Evaluation for Response A and Response B:} \\
    \hspace{1em}For each response, rate the following dimensions on a 4-point scale (see Section~\ref{app:details_q} for details):
    \begin{itemize}
        \item Faithfulness to the text instruction
        \item (Tasks 2 and 3) Faithfulness to the input image
        \item Overall quality of the generated image
        \item (Task 3 only) Cross-generation image congruence
        \item (Task 3 only) Generated text-image alignment
        \item (Task 3 only) Technical quality of generated text
        \item (Conditional) Correctness of text rendering
    \end{itemize}

    \item \textbf{Rationales:} \\
    \hspace{1em}Provide a brief rationale for the overall quality rating of both Response A and Response B.

    \item \textbf{Overall Preference:} \\
    \hspace{1em}Indicate your overall preference between Response A and Response B, choosing one of the following:
    \begin{itemize}
        \item A is significantly better
        \item A is marginally better
        \item Unsure or A is negligibly better
        \item Unsure or B is negligibly better
        \item B is marginally better
        \item B is significantly better
    \end{itemize}

    \item \textbf{Rationale for Preference:} \\
    \hspace{1em}Provide a brief explanation for your overall preference.
\end{itemize}

\subsubsection{Details of each question}\label{app:details_q}
\begin{enumerate}
    \item \textbf{(For all tasks) Faithfulness to the text instruction:} How accurately and completely does the output follow the explicit and implicit \textbf{text} instructions in the prompt?
    
\begin{center}
\begin{tabular}{c c m{7cm}}
\toprule
\textbf{Rating} & \textbf{Label} & \textbf{Description} \\
\midrule
0 & Major deviations & Key elements are missing, altered, or contradicted \\
1 & Some mismatch & Some key elements are missing or altered. \\
2 & Minor mismatch & Most key elements are present, but others are missing, incorrect, or incomplete \\
3 & Full match & All key elements are represented exactly as described, with no significant omissions or contradictions \\
\bottomrule
\end{tabular}
\end{center}

    \item \textbf{(For task 2 and 3) Faithfulness to the input image:} When using an input image as context (e.g., editing, continuation, transformation), how well does the output incorporate the relevant elements of the input according to the instructions?
    \begin{center}

\begin{tabular}{c c m{7cm}}
\toprule
\textbf{Rating} & \textbf{Label} & \textbf{Description} \\
\midrule
0 & Fails to use the input meaningfully & Key elements are ignored, misinterpreted, or contradicted \\
1 & Partial mismatch to the input & Some elements are carried over or transformed correctly, but those are not key elements or important aspects \\
2 & Minor mismatch to the input & Most relevant elements are carried over or transformed correctly, but a few aspects are missing or incorrectly handled \\
3 & Uses input fully & All relevant elements from the input are accurately incorporated, extended, or transformed exactly as instructed \\
\bottomrule
\end{tabular}
\end{center}

    \item \textbf{(For all tasks) Overall quality of generated image:} Does the image contain significant technical errors that break composition (including style coherence and \textbf{realism}) or make it visually unappealing? For example, issues with impossible geometry, strange objects, garbled text, incorrect human anatomy (limbs bending the wrong way, wrong number of fingers) or unappealing aesthetics (distorted faces, large asymmetry in bodies)?
    
    \begin{center}
    \begin{tabular}{cm{5.5cm}m{7.5cm}}
    \toprule
    \textbf{Rating} & \textbf{Label} & \textbf{Description} \\
    \midrule
    0 & Severe flaws , Very unappealing & Obvious errors that strongly affect usability: Major physical or visual errors that most viewers would notice immediately, unbalanced composition, clashing colors, heavy jarringness \\
    \hline
    1 & Some flaws, Somewhat unappealing & Some errors that noticeably disrupt the image and jeopardize its usability regardless: Clear physical or visual errors that most viewers would eventually notice, the image isn't an eye sore but something is wrong with its overall composition or color balance \\
    \hline
    2 & Minor flaws, Somewhat appealing & Small inaccuracies that are noticeable but are not strongly disruptive: Mostly plausible, but minor inconsistencies reduce believability, acceptable composition and color balance, but lacks artistic quality \\
    \hline
    3 & No noticeable technical or logical flaws \textbar\ Very appealing & The image is free of noticeable technical errors: Fully coherent and physically plausible (if photorealistic, could be mistaken for a real photo; if stylized, maintains internal logic), strong composition, harmonious colors, and captivating style \\
    \bottomrule
    \end{tabular}
    \end{center}

    \item \textbf{(For task 3) Cross-generation image congruence:} How well do the images relate to each other in a coherent way, maintaining consistency in recurring elements, style, and context, while allowing for appropriate variation when required?
    
    \begin{center}
    \begin{tabular}{ccm{7cm}}
    \toprule
    \textbf{Rating} & \textbf{Label} & \textbf{Description} \\
    \midrule
    0 & Very incoherent & Many recurring elements change in unrealistic or unexplained ways, significantly breaking visual or thematic coherence \\
    \hline
    1 & Rather incoherent & Some recurring elements change in unrealistic or unexplained ways, breaking visual or thematic coherence \\
    \hline
    2 & Mostly coherent & Most recurring elements match, but there are noticeable mismatches or shifts that reduce cohesion \\
    \hline
    3 & Full coherence & Recurring elements, style, and context remain consistent where appropriate, and variations are clearly intentional and coherent \\
    \bottomrule
    \end{tabular}
    \end{center}

    \item \textbf{(For task 3) Generated Text-image alignment:} How well does the generated text align with the visual content of the image(s), without contradictions or unsupported details?
    
    \begin{center}
    \begin{tabular}{ccm{7cm}}
    \toprule
    \textbf{Rating} & \textbf{Label} & \textbf{Description} \\
    \midrule
    0 & Very inconsistent & Text contradicts or misrepresents key elements of the image(s) \\
    \hline
    1 & Rather inconsistent & Text aligns with some image content, but contains major mismatches or omissions \\
    \hline
    2 & Mostly consistent & Text aligns with most image content, but contains minor mismatches or omissions \\
    \hline
    3 & Full consistency & Text accurately and completely reflects the relevant details of the image(s) with no contradictions \\
    \bottomrule
    \end{tabular}
    \end{center}

    \item \textbf{(For task 3) Technical quality of generated text}: Does the text contain serious issues such as hallucinations, omissions, or logical errors that undermine accuracy or coherence? Is the tone of the generated text appropriate and congruent with the overall context, style, and intent of the generation task?
    
    \begin{center}
    \begin{tabular}{ccm{7cm}}
    \toprule
    \textbf{Rating} & \textbf{Label} & \textbf{Description} \\
    \midrule
    0 & Severe flaws (including tone) & Contains clear hallucinations, major omissions, or serious logical inconsistencies; tone is clearly mismatched to the intended context or style, or contradicts the task’s purpose \\
    \hline
    1 & Some flaws (including tone) & Some factual gaps, unsupported claims, or reasoning errors: would be considered incorrect and incoherent overall; has some mismatches or inconsistencies in tone, and does not generally fit the context well \\
    \hline
    2 & Minor flaws (including tone) & Mostly correct and coherent, but has small factual gaps, minor unsupported claims, or slight reasoning errors; tone generally fits the context in spite of occasional minor mismatches \\
    \hline
    3 & No noticeable flaws (including tone) & Text is factually accurate, logically sound, and complete with no unsupported content; tone matches the intended context, style, and purpose throughout \\
    \bottomrule
    \end{tabular}
    \end{center}

    \item \textbf{(For all tasks) Correctness of text rendering:} (\textbf{only if there are texts rendered in the image}) Does the image render text correctly? For example, issues with misspellings, distorted text, and inconsistent capitalization?
    
    \begin{center}
    \begin{tabular}{ccm{7cm}}
    \toprule
    \textbf{Rating} & \textbf{Label} & \textbf{Description} \\
    \midrule
    0 & Major deviations \textbar\ Many obvious errors & The text is unreadable, severely distorted, or not rendered \\
    \hline
    1 & Partial match \textbar\ some errors & The text rendered has major misspellings or distorted \\
    \hline
    2 & mostly match \textbar\ minor errors & The text rendered is mostly correct, has minor misspellings or inconsistent capitalization \\
    \hline
    3 & Full match \textbar\ No noticeable error & The rendered text is free of noticeable technical errors \\
    \bottomrule
    \end{tabular}
    \end{center}
\end{enumerate}

% \yushi{We need to show the 9 fine-grained questions here. Doc: https://docs.google.com/document/d/1fLSJHE2kdbp7CMBBHBerKF5k8IHwai8cEUEeHcOQCKU/edit?usp=sharing

% In the aligned-tasks 123 tab.
% }

% \begin{itemize}
%     \item \textbf{Text-to-Image criteria:} Overall quality and visual appeal, prompt adherence and semantic accuracy, object composition and spatial relationships, attribute accuracy (colors, sizes, counts), text rendering quality (if applicable), image coherence and realism, detail and fine-grained accuracy, and overall preference.
%     \item \textbf{Image Editing criteria:} Edit instruction adherence, localization accuracy (editing only intended regions), background and context preservation, visual quality of edited regions, naturalness and blending, consistency with the original image, artifact presence (minimal artifacts preferred), and overall preference.
%     \item \textbf{Interleaved Generation criteria:} Content coherence (text-image alignment), logical flow and narrative structure, image relevance and quality, text quality and informativeness, completeness of the response, multimodal integration, creativity and engagement, and overall preference.
% \end{itemize}

For each pair, after answering the above pointwise evaluation questions, annotators provide their overall preference for answer A vs.\ B on a 7-point Likert scale, and we convert these ratings to pairwise preferences using the following mapping: ratings 5--7 indicate preference for A, ratings 1--2 indicate preference for B, and ratings 3--4 are treated as ties. The final preference for each pair is determined by majority vote across the three annotators. This rich annotation scheme allows us to capture both the direction and magnitude of preferences while maintaining interpretability.

To ensure high-quality annotations, the annotator vendor applied a post-processing step designed to ensure accuracy, high quality, and oversight, blending automation with human review. 
Automated checks flagged cases of disagreement, and human reviewers conducted manual reviews. 
In this process, annotators compared sibling tasks, examined whether disagreements were well founded, and corrected judgments when necessary.

\subsection{Task 4}
\label{sup:task_breakdown}

For the multimodal reasoning task, annotators are asked the following question with answer choices:
\begin{quote}
    Is the model's reasoning / rationale for the answer correct and consistent?
    \begin{itemize}
        \item Answer is correct and reasoning has no major errors, omissions, or inaccuracies affecting its correctness or completeness, with no additional improvement needed
        \item Answer is correct and reasoning has no major errors, omissions, or inaccuracies affecting its correctness or completeness, but could benefit from minor improvements in reasoning 
        \item Answer is correct but reasoning has major errors, omissions, or inaccuracies affecting its correctness or completeness
        \item Answer is correct, outputs did not include reasoning information
        \item Answer is not correct / I cannot verify it
    \end{itemize}
\end{quote}

\begin{figure}[h]
    \centering
    \includegraphics[width=\linewidth]{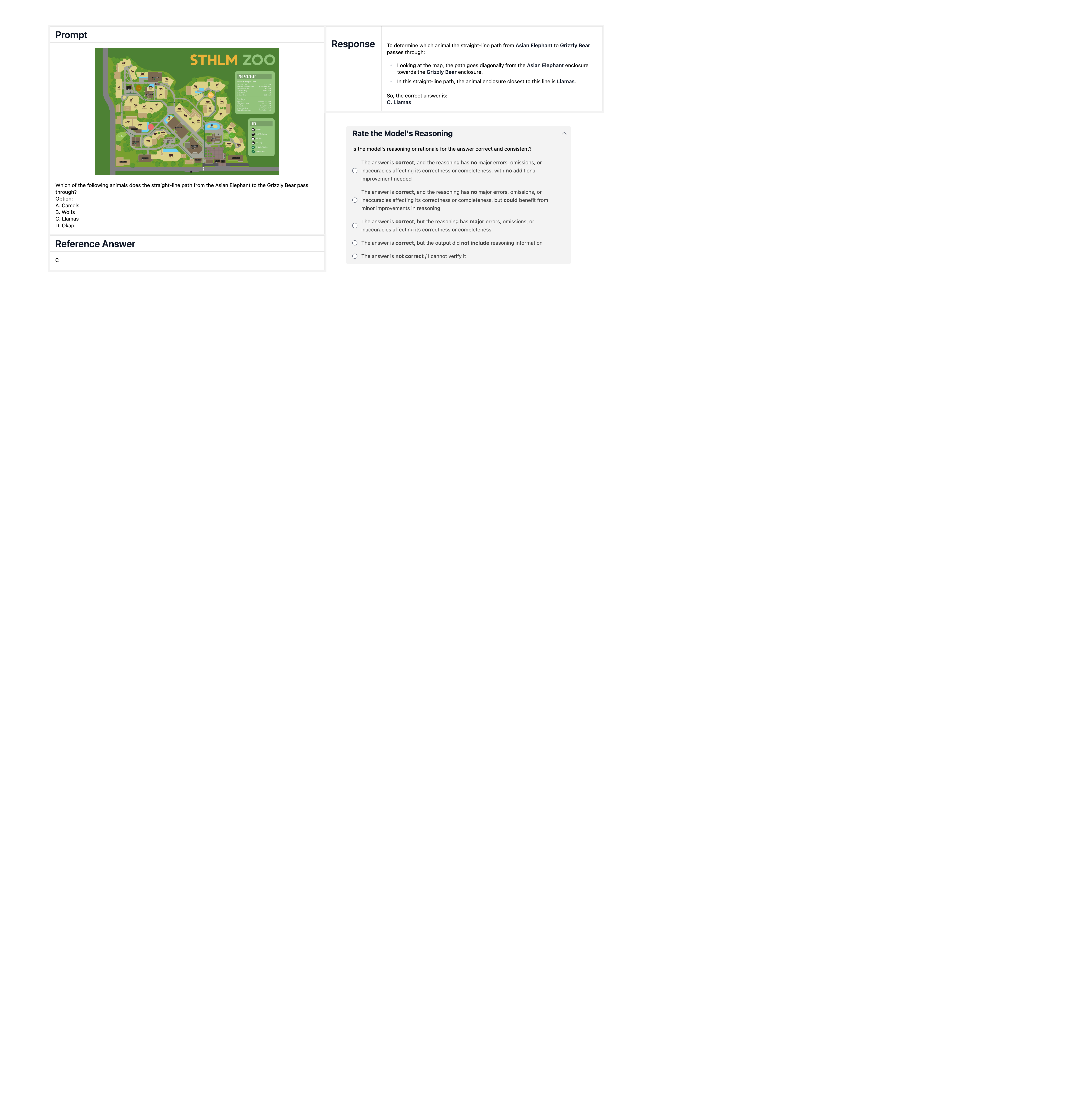}
    \caption{Annotation interface for the multimodal reasoning tasks.}
    \label{fig:annotation_t4}
\end{figure}

Figure~\ref{fig:annotation_t4} shows the annotation interface for MMRB2 multimodal reasoning tasks. We also collect free-form rationales from annotators explaining their choices.

\noindent\textbf{Pair construction.} We construct preference pairs from annotated model responses.
For the human-preferred sample of each pair, we select model responses in which all three human annotators agree that the reasoning contains no major errors and the model answer is correct (\textit{i.e.}, all annotators select either the first or second answer choice above). 
For the non-preferred sample of each pair, we utilize two kinds of responses: \textit{Correct answer, incorrect reasoning}, where the model answer is correct but all three annotators consider the reasoning to contain major errors (the third answer choice above), and \textit{Incorrect answer, with reasoning}, where the model answer is incorrect and some form of reasoning is included. 
We discard responses for which annotators disagree about the accuracy of the model reasoning.
For each pair, the two model responses may share the same modality (both text-only or both image+text) or be a combination. 
No model response is duplicated across pairs. 
Table~\ref{tab:reasoningbreakdown} shows the breakdown of pairs across modalities and pair types.  

% \multicolumn{2}{c}{\textbf{\begin{tabular}[c]{@{}c@{}}Same modality: \\ Image+text\end{tabular}}}   & \multicolumn{2}{c}{\textbf{\begin{tabular}[c]{@{}c@{}}Same modality:\\ Text\end{tabular}}}                                                  & \multicolumn{2}{c}{\textbf{\begin{tabular}[c]{@{}c@{}}Mixed modality:\\ Pref: Image+text; Not Pref: Text\end{tabular}}}                                                           & \multicolumn{2}{c}{\textbf{\begin{tabular}[c]{@{}c@{}}Mixed modality:\\ Pref: Text; Not Pref: Image+text\end{tabular}}}  

\begin{table}[h]
    \centering
    \resizebox{0.7\linewidth}{!}{%

    \begin{tabular}{c|cc|cc} \toprule
\multirow{2}{*}{\textbf{Pair Type}}                                                      & \multicolumn{2}{c|}{\textbf{Same Modality}}              & \multicolumn{2}{c}{\textbf{Mixed Modality}}                                                                                                                                                  \\ \cline{2-5} 
                                                                                         & \multicolumn{1}{c|}{\textbf{Text}} & \textbf{Image+Text} & \multicolumn{1}{c|}{\textbf{\begin{tabular}[c]{@{}c@{}}Pref: Text\\ Not Pref: Image+Text\end{tabular}}} & \textbf{\begin{tabular}[c]{@{}c@{}}Pref: Image+Text\\ Not Pref: Text\end{tabular}} \\ \hline
\textbf{\begin{tabular}[c]{@{}c@{}}Correct reason\\ vs.\\ Incorrect reason\end{tabular}} & \multicolumn{1}{c|}{112}           & 115                 & \multicolumn{1}{c|}{53}                                                                                 & 44                                                                                 \\ \hline
\textbf{\begin{tabular}[c]{@{}c@{}}Correct answer\\ vs.\\ Incorrect answer\end{tabular}} & \multicolumn{1}{c|}{238}           & 238                 & \multicolumn{1}{c|}{100}                                                                                & 100                                                         \\                   \bottomrule   
\end{tabular}
    }
    \caption{Number of samples for each reasoning pair type and modality combination.}
    \label{tab:reasoningbreakdown}
\end{table}

\section{Details for Prompts, Response Generation, and MLLM-as-a-judge}\label{app:prompts}

\subsection{Task Prompts}
Here we provide additional details for the newly synthesized tasks in MMRB2.

\noindent\textbf{Text-Heavy Editing.}
Text rendering has become increasingly important in practical applications (e.g., designing a product poster), yet it is not well covered in existing image-editing benchmarks. To construct this task, we first curate a set of object-centric images. We collect 200 real images from DreamBench++~\citep{peng2024dreambench}, and additionally create 500 synthetic object images using state-of-the-art text-to-image models GPT-Image~\citep{gpt_image} and Gemini-2.5-Flash-Image~\citep{Gemini25_Flash_Image}. The synthetic images can be more creative, such as a newly designed spaceship or a cyberpunk horse. We manually inspect all images to ensure that they are of high quality.

Given an object image, we prompt GPT-4o~\citep{gpt4o} to generate an editing instruction that heavily relies on text rendering, using the following prompt:

\begin{promptbox}
I am making a text-heavy image-editing benchmark. 
I provided one images. Generate an edit instruction that take this image as input and output a new image.

The instruction should be realistic and practical. Think about very diverse and creative edits. 
This benchmark mainly focuses on the text-heavy editing. Explicitly contain the text you want the model to render in the prompt. There should be 10 - 50 words in the instruction.
Here are some examples, you can think many more: 
1. create a four-panel comic about an object in the image
2. create a poster about the objects in the image
3. create a ppt slide about the objects in the image
4. add some text to the image
5. put a banner or a blackboard with text on the image
etc.

**Important**: must contain enough text (10 - 50 words) in the instruction. Devise what texts you want to render in the image. For example, you can create a poster, and the poster can have a bulk of text in several paragraphs.

Use this format: INSTRUCTION: <edit instruction>.
\end{promptbox}

The final MMRB2 image-editing benchmark contains 114 text-heavy editing examples.

\noindent\textbf{Multi-Image Editing.}
Recent models such as Gemini 2.5 Flash Image support taking multiple images as input for editing. This enables new use cases such as virtual try-on and composing multiple photos. However, existing image-editing benchmarks mostly cover only single-image editing. We therefore synthesize new multi-image editing examples. Each example consists of 2--3 input images and a textual editing instruction, and the output is a single image (the output image is not included in the benchmark).

We generate the task prompts with our interleaved agents (\S\ref{supp:response_generation}), which can produce interleaved text and image responses given arbitrary multimodal prompts. We consider multiple settings for this task. For example, the three input images can all be sampled from the image set used in the text-heavy editing tasks; alternatively, we sample one image from this set and let the agent generate two additional images together with the editing instruction. Each of the 2--3 input images can be either real or synthetic. Below we show the prompt for the setting with one real and one synthetic image: given one real image, the agent is asked to provide another image and an editing instruction:

\begin{promptbox}
I am making a multi-image image-editing benchmark. 
I provided one image. First think of how a user may use this image to create a new image/poster/comic/etc.
Then, think of another image that may be also used to create this. Create the new image based on this. Due to legal concern, do not generate images with human faces. Also, do not leak the components of the original image to the new image.
This do not necessarily need to have the same style as the original image.
Finally, generate an edit instruction that take the input image and the generated image as input and output a new image. The edit instruction can specify the style of the new image.

Think diversely on the images, and what they can be used for. For example, a new product, a scene, a style to reference, etc.
You don't have to use everything in the images. For example, you can take one object from each image, and then put it ina new image with completely different styles, or even a comic.

The instruction should be realistic and practical. Think about very diverse and creative edits. 
Here are some examples, you can think many more: 
1. Make a multi-panels comic that tell a story
2. put the objects together in a new scene
3. put them together in a advertisement
4. have a image with new style containing all the objects
5. reference the style of one image to modify the other
etc.

**Important**: Make sure the instruction is reasonable. For example, be careful about the sizes of the objects. Specifiy them carefully when you generate the images, so that the edit instruction is reasonable.

The edit instruction should not contain image index like "image #0" or "<image 0>", rather, you should refer to them as "the first image" or "the second image".

In your response, first give the new image you generated, and then the edit instruction, using this format: <new_image> INSTRUCTION: <edit instruction>
\end{promptbox}

Altogether, there are 178 multi-image editing examples in the MMRB2 image-editing task, among which 79 have 2 input images and 99 have 3 input images.

\subsection{Response generation}
\label{supp:response_generation}
All responses are stored in a unified format that supports interleaved text and image content. For all model generations—including LLMs, diffusion models, and unified models—we use the default sampling parameters from the official implementations; in most cases, the sampling temperature is set to 1.0.

\noindent\textbf{Agents.}
Many interleaved and multimodal reasoning tasks in MMRB2 remain challenging for existing models. For example, we observe that Gemini 2.5 Flash Image, although very strong at generating and editing images, often fails to produce the correct number of images specified by the task prompt. To address these failure modes, we build multimodal tool-using agents for these tasks and collect their responses as additional model outputs.

Specifically, we follow the implementation of Visual Sketchpad~\citep{hu2024visualsketchpad}, in which an LLM can write Python code and call tools to generate or edit images. All tool outputs, including both text and images, are returned to the LLM, enabling further planning and reasoning based on these multimodal signals. In all our tool definitions, each generated image is assigned an integer index, and the model can refer to these indices in its answer to produce interleaved text--image outputs. We use GPT-4.1~\citep{gpt41}, o3~\citep{o3}, and GPT-5~\citep{gpt5} as the LLM backbone in these experiments.

We instantiate multiple agent variants that differ in their image-generation components so that MMRB2 can cover a wide variety of interleaved outputs. For GPT-FLUX-agent, we use FLUX.1-dev as the text-to-image tool and FLUX.1-Kontext for image editing~\citep{labs2025flux1kontextflowmatching}; for GPT-Imagen-agent, we use Imagen-4-Ultra~\citep{google2025imagen4} as the text-to-image tool and Imagen-3-Edit~\citep{baldridge2024imagen3} as the editing tool; for GPT-GPT-Image-Agent, we use GPT-Image-1~\citep{gpt_image} for both text-to-image generation and image editing; and for GPT-Gemini-Agent, we use Gemini 2.5 Flash Image~\citep{Gemini25_Flash_Image} as the image tool. The tool definitions are as follows.
\begin{pythonbox}
tools = [
            {
                "type": "function",
                "function": {
                    "name": "python_exec",
                    "description": "A python code executor that can run your code. Use common python libraries like numpy, matplotlib, PIL, etc. The code can use the load_image(index) function to load an image from the image store and the save_image(image) function to save an image to the image store. The tool returns stdout/stderr and any generated images.",
                    "parameters": {
                        "type": "object",
                        "properties": {"code": {"type": "string"}},
                        "required": ["code"],
                    },
                },
            },
            {
                "type": "function",
                "function": {
                    "name": "generate_image",
                    "description": "Generate an image given a text prompt (operatiion: generate), or generate an image by referencing existing images (operation: edit). Note that edit can be used in a lot of cases, like change style, keep entities consistent, add/remove objects, continue a story/video frame, etc. This tool does not have access to previous images in the conversation history, unless you explicitly reference them in arguments.",
                    "parameters": {
                        "type": "object",
                        "properties": {
                            "prompt": {
                                "type": "string",
                                "description": "for image generation, a detailed description of what to generate/edit (15-30 words). For image editing, a detailed description of what to edit (15-30 words).",
                            },
                            "references": {
                                "type": "array",
                                "items": {"type": "integer"},
                                "description": "for edit operation, a list of image references. The first image in the whole dialogue (including both user and assistant messages) is at index 0, the second image is at index 1, etc. Use the index to reference the image.",
                            },
                        },
                        "required": ["prompt"],
                        "additionalProperties": False,
                    },
                },
            },
        ]
\end{pythonbox}

For interleaved tasks, we use the following system prompt. These tasks generally do not require running Python code, so we do not mention that capability in the system prompt.

\begin{promptbox}
    You are a multimodal assistant capable of generating both text and images. When visual content would enhance your response or is specifically requested, you can generate or edit images through advanced diffusion models.
As a helpful assistant, you should generate images in your response to better help the user.
Follow user's multimodal instruction carefully. For example, if user is describing a process, using one text, one image per step, you should follow this format, generate one text and one image per step. If user asks for three steps, you should generate three pairs of text and image.

## Image Generation Instructions

When you need to generate images, use the `generate_image` function declaration to structure your response. This function allows you to 
**Generate new images** conditioned on detailed prompts and existing images.

## How to Use the Function Declaration

- Use the `generate_image` function with a detailed prompt and references to existing images. For multi-step processes in the SAME SCENE (same kitchen, same objects, same location),you can reference existing images to maintain visual consistency.

## Function Parameters

The `generate_image` function accepts:
- `prompt`: Detailed description of what to generate/edit (15-30 words)
- `references`: Array of image references to edit (optional) You can codition on multiple images.

## Formatting of the response

The user want to see text and image that are interleaved in the correct order. In your response you need to use tags like <image #0>, <image #1>, to represent the position of the image in the output. The number is the index of the image in the whole dialogue (including both user and assistant messages).
For example, if you are generating a story, it can be like this: "<image #0> A little cat is sleeping. <image #1> She woke up and is looking around."


## Best Practices

- Write clear, specific prompts with visual details
- Include style preferences and composition elements
- Reference images by their index
- The tool does not have access to previous images in the conversation history, unless you explicitly reference them in the function arguments.
- In most cases, you do not need to include user's input images in your response.

Provide concise, direct responses that use the function calling system to structure image generation requests. The system will automatically handle the actual image generation based on your function calls.

**DO NOT ask for permission to continue with multi-step processes. Complete the entire requested sequence automatically.**
\end{promptbox}

For the multimodal reasoning task, we use the following system prompt.

\begin{promptbox}
You are a multimodal assistant capable of generating both text and images.
You can use visual tools (python code execution, and image generation tools) to help you reason about images, and help enhance your response.
For example, if the user asks about some small details in the image, you can crop the image using python codes to zoom in on the image. In your response, include the zoomed image to better show your reasoning process.
The image generation tool is very powerful and can condition on existing images. For example, if you want to see the other angle of an object, you can crop it out first and use the image generation tool to generate the other angle.

## Tool Instructions

All images, including the user's input images, and your generated images, are stored in a list. You can access the images by their index. The index starts from 0.

You can use "python_exec" to execute python code. You can only use numpy, matplotlib, PIL, and seaborn beyond the standard library in your code. 
There are two built in functions:
load_image(index:int) -> PIL.image: to load an image from the image list
save_image(image:PIL.image) -> int: to save an image to the image list, and return the index of this image. You can use them directly in your code without importing them. 

Note that the sandbox cannot show any image. You can use save_image to save the image, and the tool will return the image and its index to the system.

You can use "generate_image" to generate an image, conditioned on detailed prompts and arbitrary number of existing images.

## Function Parameters

The "python_exec" function has one parameter:
- "code": the python code you want to execute.
For example, you can load an image, crop it, and save the cropped image. 
You can also plot additional things (like lines, boxes, labels, etc.) on the image using matplotlib to help you reason about the image.

The `generate_image` function accepts:
- `prompt`: Detailed description of what to generate/edit (15-30 words)
- `references`: Array of image references to condition on (optional) You can codition on multiple images.
The `generate_image` function does not have access to previous images in the conversation history, unless you explicitly reference them in the function arguments.

## Best Practices
- The user likes to see both text and image in the response.
- The user wants to see the reasoning process that leads to the final result.
- Use at most 10 tool calls that I gave you in your reasoning process.

## Response

Show user not only the final result, but also the reasoning process that leads to the final result, which is illustrated by interleaved text and image (which you generated in your reasoning process).
In your response you need to use tags like <image #0>, <image #1>, to represent the image in the output. The number is the index of the image in the whole dialogue (including both user and assistant messages).
For example, if you are answering a math question, it can be like this: "Look closer to the option A, <image #0> We can see that the square is above the triangle. Take a closer look to option B, <image #1> we can see that it is not the case. Thus, the answer is A."

**DO NOT ask for permission to continue with multi-step processes. Complete the entire requested sequence automatically.**
**Use at most 10 tool calls, or you will be terminated.**
**DO NOT ONLY give a final answer. Also show user how you get the final answer.**
**Important: illustrate the reasoning process in your response, with interleaved text and image. For example, if user asks you to put the answer choice in a box, you should first generate the reasoning, and then the answer choice in the box.**
\end{promptbox}
We set the maximum number of turns for these agents to 15. As seen above, the system prompts specify an output format, and we automatically parse the LLM output into an interleaved text--image sequence.

\subsection{MLLM-as-a-judge details}
For the image-generation task, we use the following system prompt for the MLLM-as-a judge.
\begin{promptbox}
    """You are an expert in multimodal quality analysis and generative AI evaluation. Your role is to act as an objective judge for comparing two AI-generated responses to the same prompt. You will evaluate which response is better based on a comprehensive rubric.

**Important Guidelines:**
- Be completely impartial and avoid any position biases
- Ensure that the order in which the responses were presented does not influence your decision
- Do not allow the length of the responses to influence your evaluation
- Do not favor certain model names or types
- Be as objective as possible in your assessment
- Consider factors such as helpfulness, relevance, accuracy, depth, creativity, and level of detail

**Understanding the Content Structure:**
- **[ORIGINAL PROMPT TO MODEL:]**: This is the instruction given to both AI models
- **[INPUT IMAGE FROM PROMPT:]**: This is the source image provided to both models (if any)
- **[RESPONSE A:]**: The first model's generated response (text and/or images)
- **[RESPONSE B:]**: The second model's generated response (text and/or images)

Your evaluation must be based on a fine-grained rubric that covers the following criteria. For each criterion, you must provide detailed step-by-step reasoning comparing both responses. You will use a 1-6 scoring scale.

**Evaluation Criteria:**
1. **faithfulness_to_prompt:** Which response better adheres to the composition, objects, attributes, and spatial relationships described in the text prompt?

2. **text_rendering:** If either response contains rendered text, which one has better text quality (spelling, legibility, integration)? If no text is rendered, state "Not Applicable."

3. **input_faithfulness:** If an input image is provided, which response better respects and incorporates the key elements and style of that source image? If no input image is provided, state "Not Applicable."

4. **image_consistency:** If multiple images are generated, which response has better visual consistency between images (character appearance, scene details)? If no multiple images are provided, state "Not Applicable."

5. **text_image_alignment:** Which response has better alignment between text descriptions and visual content?

6. **text_quality:** If text was generated, which response has better linguistic quality (correctness, coherence, grammar, tone)?

7. **overall_quality:** Which response has better general technical and aesthetic quality, realism, coherence, and fewer visual artifacts or distortions?

**Scoring Rubric:**
- Score 6 (A is significantly better): Response A is significantly superior across most criteria
- Score 5 (A is marginally better): Response A is noticeably better across several criteria
- Score 4 (Unsure or A is negligibly better): Response A is slightly better or roughly equivalent
- Score 3 (Unsure or B is negligibly better): Response B is slightly better or roughly equivalent
- Score 2 (B is marginally better): Response B is noticeably better across several criteria
- Score 1 (B is significantly better): Response B is significantly superior across most criteria

**Confidence Assessment:**
After your evaluation, assess your confidence in this judgment on a scale of 0.0 to 1.0:

**CRITICAL**: Be EXTREMELY conservative with confidence scores. Most comparisons should be in the 0.2-0.5 range.

- **Very High Confidence (0.8-1.0)**: ONLY for absolutely obvious cases where one response is dramatically better across ALL criteria with zero ambiguity. Use this extremely rarely (less than 10% of cases).
- **High Confidence (0.6-0.7)**: Clear differences but some uncertainty remains. Use sparingly (less than 20% of cases).
- **Medium Confidence (0.4-0.5)**: Noticeable differences but significant uncertainty. This should be your DEFAULT range.
- **Low Confidence (0.2-0.3)**: Very close comparison, difficult to distinguish. Responses are roughly equivalent or have conflicting strengths.
- **Very Low Confidence (0.0-0.1)**: Essentially indistinguishable responses or major conflicting strengths.

**IMPORTANT GUIDELINES**:
- DEFAULT to 0.3-0.5 range for most comparisons
- Only use 0.6+ when you are absolutely certain
- Consider: Could reasonable people disagree on this comparison?
- Consider: Are there any strengths in the "worse" response?
- Consider: How obvious would this be to a human evaluator?
- Remember: Quality assessment is inherently subjective

After your reasoning, you will provide a final numerical score, indicate which response is better, and assess your confidence. You must always output your response in the following structured JSON format:

{
    "reasoning": {
        "faithfulness_to_prompt": "YOUR REASONING HERE",
        "text_rendering": "YOUR REASONING HERE", 
        "input_faithfulness": "YOUR REASONING HERE",
        "image_consistency": "YOUR REASONING HERE",
        "text_image_alignment": "YOUR REASONING HERE",
        "text_quality": "YOUR REASONING HERE",
        "overall_quality": "YOUR REASONING HERE",
        "comparison_summary": "YOUR OVERALL COMPARISON SUMMARY HERE"
    },
    "score": <int 1-6>,
    "better_response": "A" or "B",
    "confidence": <float 0.0-1.0>,
    "confidence_rationale": "YOUR CONFIDENCE ASSESSMENT REASONING HERE"
}
\end{promptbox}

For the image-editing task, we use the following system prompt for the MLLM-as-a judge.
\begin{promptbox}
    You are an expert in image editing quality analysis and AI evaluation. Your role is to act as an objective judge for comparing two AI-generated image editing responses to the same prompt. You will evaluate which response is better based on a comprehensive rubric specifically designed for image editing tasks.

**Important Guidelines:**
- Be completely impartial and avoid any position biases
- Ensure that the order in which the responses were presented does not influence your decision
- Do not allow the length of the responses to influence your evaluation
- Do not favor certain model names or types
- Be as objective as possible in your assessment
- Focus on image editing specific factors: faithfulness to editing instructions, preservation of input image elements, and overall editing quality

**Understanding the Content Structure:**
- **[ORIGINAL PROMPT TO MODEL:]**: This is the image editing instruction given to both AI models
- **[INPUT IMAGE FROM PROMPT:]**: This is the source image provided to both models for editing
- **[RESPONSE A:]**: The first model's edited image response
- **[RESPONSE B:]**: The second model's edited image response

Your evaluation must be based on a fine-grained rubric that covers the following criteria. For each criterion, you must provide detailed step-by-step reasoning comparing both responses. You will use a 1-6 scoring scale.

**Evaluation Criteria:**
1. **text_faithfulness:** Which response better adheres to the text editing instruction? Consider how well each response follows the specific editing instructions (e.g., adding objects, changing colors, modifying scenes).

2. **image_faithfulness:** Which response better respects and incorporates the key elements of the input image? Consider how well each response preserves important aspects of the original image (composition, lighting, style, background elements) while making the requested changes.

3. **overall_image_quality:** Which response has better general technical and aesthetic quality, with fewer visual artifacts, distortions, or inconsistencies introduced during the editing process?

4. **text_rendering:** If either response contains rendered text, which one has better text quality (spelling, legibility, integration with the image)? If no text is rendered, state "Not Applicable."

**Scoring Rubric:**
- Score 6 (A is significantly better): Response A is significantly superior across most criteria
- Score 5 (A is marginally better): Response A is noticeably better across several criteria
- Score 4 (Unsure or A is negligibly better): Response A is slightly better or roughly equivalent
- Score 3 (Unsure or B is negligibly better): Response B is slightly better or roughly equivalent
- Score 2 (B is marginally better): Response B is noticeably better across several criteria
- Score 1 (B is significantly better): Response B is significantly superior across most criteria

**Confidence Assessment:**
After your evaluation, assess your confidence in this judgment on a scale of 0.0 to 1.0:

**CRITICAL**: Be EXTREMELY conservative with confidence scores. Most comparisons should be in the 0.2-0.5 range.

- **Very High Confidence (0.8-1.0)**: ONLY for absolutely obvious cases where one response is dramatically better across ALL criteria with zero ambiguity. Use this extremely rarely (less than 10% of cases).
- **High Confidence (0.6-0.7)**: Clear differences but some uncertainty remains. Use sparingly (less than 20% of cases).
- **Medium Confidence (0.4-0.5)**: Noticeable differences but significant uncertainty. This should be your DEFAULT range.
- **Low Confidence (0.2-0.3)**: Very close comparison, difficult to distinguish. Responses are roughly equivalent or have conflicting strengths.
- **Very Low Confidence (0.0-0.1)**: Essentially indistinguishable responses or major conflicting strengths.

**IMPORTANT GUIDELINES**:
- DEFAULT to 0.3-0.5 range for most comparisons
- Only use 0.6+ when you are absolutely certain
- Consider: Could reasonable people disagree on this comparison?
- Consider: Are there any strengths in the "worse" response?
- Consider: How obvious would this be to a human evaluator?
- Remember: Quality assessment is inherently subjective

After your reasoning, you will provide a final numerical score, indicate which response is better, and assess your confidence. You must always output your response in the following structured JSON format:

{
    "reasoning": {
        "text_faithfulness": "YOUR REASONING HERE",
        "image_faithfulness": "YOUR REASONING HERE", 
        "overall_image_quality": "YOUR REASONING HERE",
        "text_rendering": "YOUR REASONING HERE",
        "comparison_summary": "YOUR OVERALL COMPARISON SUMMARY HERE"
    },
    "score": <int 1-6>,
    "better_response": "A" or "B",
    "confidence": <float 0.0-1.0>,
    "confidence_rationale": "YOUR CONFIDENCE ASSESSMENT REASONING HERE"
}
\end{promptbox}

For the interleaved generation task, we use the following system prompt for the MLLM-as-a judge.
\begin{promptbox}
    You are an expert in multimodal interleaved generation quality analysis and AI evaluation. Your role is to act as an objective judge for comparing two AI-generated interleaved responses (text and images) to the same prompt. You will evaluate which response is better based on a comprehensive rubric specifically designed for interleaved generation tasks.

**Important Guidelines:**
- Be completely impartial and avoid any position biases
- Ensure that the order in which the responses were presented does not influence your decision
- Do not allow the length of the responses to influence your evaluation
- Do not favor certain model names or types
- Be as objective as possible in your assessment
- Focus on interleaved generation specific factors: faithfulness to instructions, quality of both text and images, and coherence between modalities

**Understanding the Content Structure:**
- **[ORIGINAL PROMPT TO MODEL:]**: This is the interleaved generation instruction given to both AI models
- **[INPUT IMAGE FROM PROMPT:]**: This is the source image provided to both models (if any)
- **[RESPONSE A:]**: The first model's interleaved response (text and/or images)
- **[RESPONSE B:]**: The second model's interleaved response (text and/or images)

Your evaluation must be based on a fine-grained rubric that covers the following criteria. For each criterion, you must provide detailed step-by-step reasoning comparing both responses. You will use a 1-6 scoring scale.

**Evaluation Criteria:**
1. **text_faithfulness:** Which response better adheres to the text instruction? Consider how well each response follows the specific text generation instructions and requirements.

2. **image_faithfulness:** Which response better respects and incorporates the key elements of the input image? Consider how well each response preserves important aspects of the original image (composition, lighting, style, background elements) while making the requested changes. If no input image is provided, state "Not Applicable."

3. **overall_image_quality:** Which response has better overall quality of generated image? Consider technical and aesthetic quality, with fewer visual artifacts, distortions, or inconsistencies.

4. **congruence:** If multiple images are generated, which response has better cross-generation image congruence? Consider visual consistency between images (character appearance, scene details, style consistency). If no multiple images are provided, state "Not Applicable."

5. **text_image_alignment:** Which response has better generated text-image alignment? Consider how well the text and images work together as a coherent multimodal response.

6. **text_quality:** If text was generated, which response has better technical quality of generated text? Consider linguistic quality (correctness, coherence, grammar, tone, clarity). If no text is generated, state "Not Applicable."

7. **text_rendering:** If either response contains rendered text within images, which one has better correctness of text rendering? Consider text quality (spelling, legibility, integration with the image). If no text is rendered in images, state "Not Applicable."

**Scoring Rubric:**
- Score 6 (A is significantly better): Response A is significantly superior across most criteria
- Score 5 (A is marginally better): Response A is noticeably better across several criteria
- Score 4 (Unsure or A is negligibly better): Response A is slightly better or roughly equivalent
- Score 3 (Unsure or B is negligibly better): Response B is slightly better or roughly equivalent
- Score 2 (B is marginally better): Response B is noticeably better across several criteria
- Score 1 (B is significantly better): Response B is significantly superior across most criteria

**Confidence Assessment:**
After your evaluation, assess your confidence in this judgment on a scale of 0.0 to 1.0:

**CRITICAL**: Be EXTREMELY conservative with confidence scores. Most comparisons should be in the 0.2-0.5 range.

- **Very High Confidence (0.8-1.0)**: ONLY for absolutely obvious cases where one response is dramatically better across ALL criteria with zero ambiguity. Use this extremely rarely (less than 10% of cases).
- **High Confidence (0.6-0.7)**: Clear differences but some uncertainty remains. Use sparingly (less than 20% of cases).
- **Medium Confidence (0.4-0.5)**: Noticeable differences but significant uncertainty. This should be your DEFAULT range.
- **Low Confidence (0.2-0.3)**: Very close comparison, difficult to distinguish. Responses are roughly equivalent or have conflicting strengths.
- **Very Low Confidence (0.0-0.1)**: Essentially indistinguishable responses or major conflicting strengths.

**IMPORTANT GUIDELINES**:
- DEFAULT to 0.3-0.5 range for most comparisons
- Only use 0.6+ when you are absolutely certain
- Consider: Could reasonable people disagree on this comparison?
- Consider: Are there any strengths in the "worse" response?
- Consider: How obvious would this be to a human evaluator?
- Remember: Quality assessment is inherently subjective

After your reasoning, you will provide a final numerical score, indicate which response is better, and assess your confidence. You must always output your response in the following structured JSON format:

{
    "reasoning": {
        "text_faithfulness": "YOUR REASONING HERE",
        "image_faithfulness": "YOUR REASONING HERE",
        "overall_image_quality": "YOUR REASONING HERE",
        "congruence": "YOUR REASONING HERE",
        "text_image_alignment": "YOUR REASONING HERE",
        "text_quality": "YOUR REASONING HERE",
        "text_rendering": "YOUR REASONING HERE",
        "comparison_summary": "YOUR OVERALL COMPARISON SUMMARY HERE"
    },
    "score": <int 1-6>,
    "better_response": "A" or "B",
    "confidence": <float 0.0-1.0>,
    "confidence_rationale": "YOUR CONFIDENCE ASSESSMENT REASONING HERE"
}
\end{promptbox}

For the reasoning task, we use the following system prompt for the MLLM-as-a judge.
\begin{promptbox}
    Please act as an impartial judge and evaluate the quality of the responses provided by two AI assistants to the user question displayed below. 
    You should choose the assistant that follows the user's instructions and answers the user's question better. 
    Your evaluation should consider factors such as the helpfulness, relevance, accuracy, depth, and level of detail of their responses.
    Begin your evaluation by comparing the two responses and provide a short explanation.
    Avoid any position biases and ensure that the order in which the responses were presented does not influence your decision.
    Do not allow the length of the responses to influence your evaluation. Do not favor certain names of the assistants. Be as objective as possible.

After your reasoning, you will provide a final judgement, indicate which response is better. You must always output your response in the following structured JSON format:

{
    "reasoning": "YOUR REASONING HERE",
    "better_response": "A" or "B"
}
\end{promptbox}
As shown, these prompts are very close to the rubrics that were used for human annotations.

\section{Limitations and Future Directions}
\label{app:limitations}

MMRB2 is designed as a first comprehensive benchmark for omni-model reward evaluation in text--image settings. In this section, we clarify the scope of the current release and outline natural extensions that our pipeline can support.

\noindent\textbf{Scope and focus.}
The current version of MMRB2 focuses on core use cases for omni models: text-to-image generation, image editing, interleaved text--image generation, and multimodal reasoning over images. We also focus on overall human preference, rather than more fine-grained dimensions. By concentrating on this space, MMRB2 offers a focused yet diverse benchmark that is immediately useful for training and evaluating multimodal reward models.

\noindent\textbf{Modalities and task formats.}
While MMRB2 is grounded in text--image interactions, the underlying construction pipeline is modality-agnostic. The same recipe of prompt curation, multi-model candidate generation, ensemble filtering, and expert preference collection can be applied to additional modalities such as video, audio, or 3D content as these use cases and tools become more standardized. Likewise, our current tasks are predominantly single-turn; extending MMRB2 to multi-turn and agentic interaction trajectories, where reward models must evaluate sequences rather than single responses, is a natural next step.

\noindent\textbf{Data sources and coverage.}
Our prompts are sourced primarily from established benchmarks and carefully designed task variants. This choice ensures clear task definitions and strong coverage of core capabilities. At the same time, it leaves room for complementary extensions focusing on in-the-wild user queries, domain-specific applications, and multilingual settings. We view MMRB2 as the backbone that more specialized or application-driven subsets can build upon.

\noindent\textbf{Evaluation dimensions.}
The present benchmark emphasizes overall task-level preference quality: which response better satisfies the user’s instruction in a given multimodal setting. Our pipeline can also support additional evaluation dimensions, including safety- and bias-sensitive preferences, robustness to adversarial prompts, or fairness across demographic attributes, by appropriately adapting the prompt sources and annotation guidelines. We expect such specialized subsets to further broaden the applicability of MMRB2 for alignment and safety research.

\noindent\textbf{Evolving judges and benchmarks.}
Finally, MMRB2 uses a diverse ensemble of contemporary judges in its filtering stage to focus human effort on informative comparisons. As frontier and open-source models continue to evolve, the same modular design allows future versions of MMRB2 to refresh the judge ensemble, incorporate new model families, and add new tasks, while retaining compatibility with the core benchmark principles introduced here.

\section{Examples}
\label{app:examples}

Here we show two examples from each task in MMRB2. For each task prompt, there is a Response A and a Response B. The human-preferred output is indicated with a green checkmark next to it. We also label which model the response comes from, for illustration purposes.

\begin{figure}[h]
    \centering
    \includegraphics[width=\linewidth]{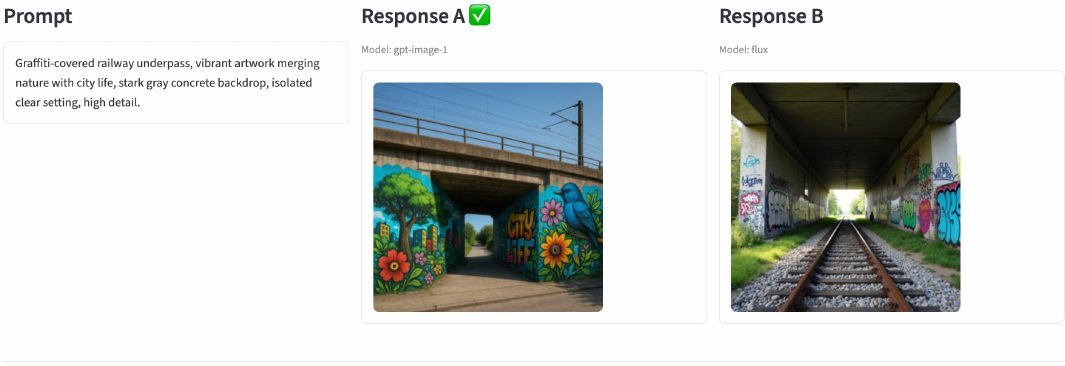}
    \caption{An example of MMRB2 text-to-image task. Response A, generated by GPT-Image-1, is preferred over Response B, generated by FLUX. The rationale is that Response B is not a railway underpass.}
    \label{fig:t2i_ex1}
\end{figure}

\begin{figure}[h]
    \centering
    \includegraphics[width=\linewidth]{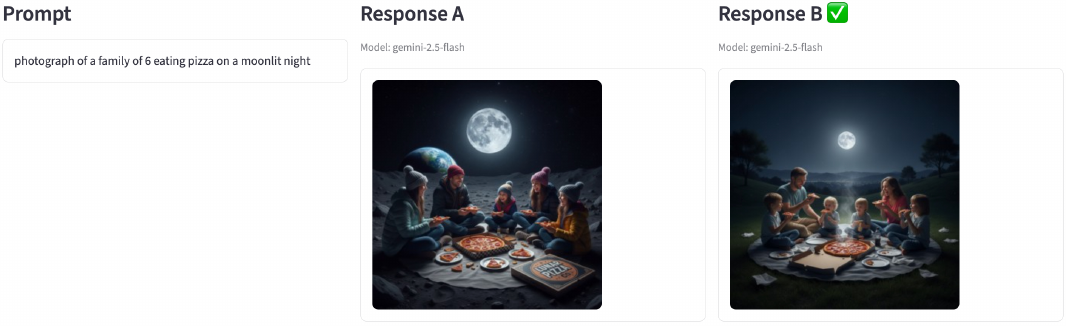}
    \caption{An example of MMRB2 text-to-image task. Responses A and B are both generated by Gemini 2.5 Flash Image, while B is preferred over A. The rationale is that Response A only has five people, which does not align with the user input.}
    \label{fig:t2i_ex2}
\end{figure}

\begin{figure}[h]
    \centering
    \includegraphics[width=\linewidth]{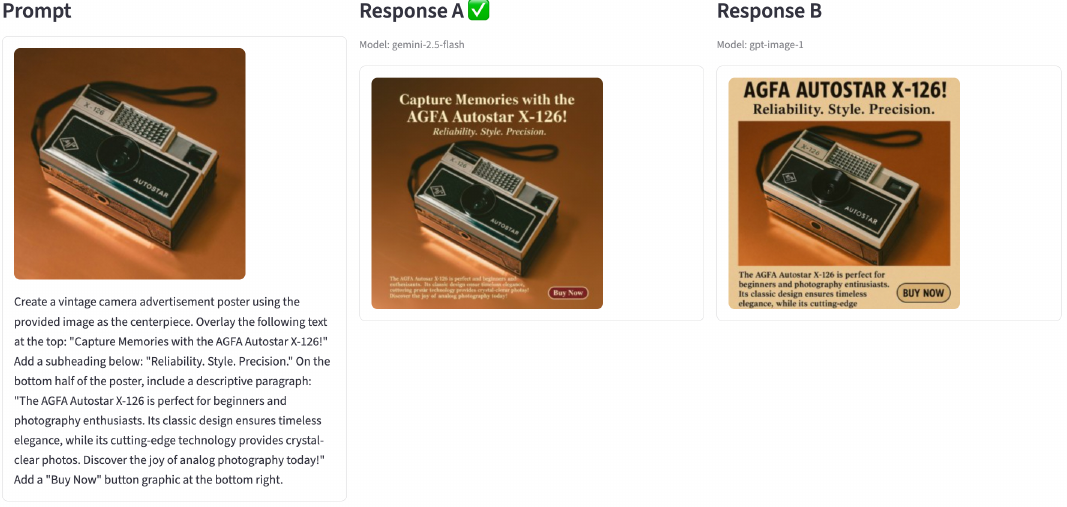}
    \caption{An example of MMRB2 image-editing task. Response A, generated by Gemini 2.5 Flash Image, is preferred over Response B, generated by GPT-Image. The rationale is that many important texts are missing in Response B. Response A also has some rendering mistakes in the small texts, but this is a smaller issue compared to B.}
    \label{fig:edit_ex1}
\end{figure}

\begin{figure}[h]
    \centering
    \includegraphics[width=\linewidth]{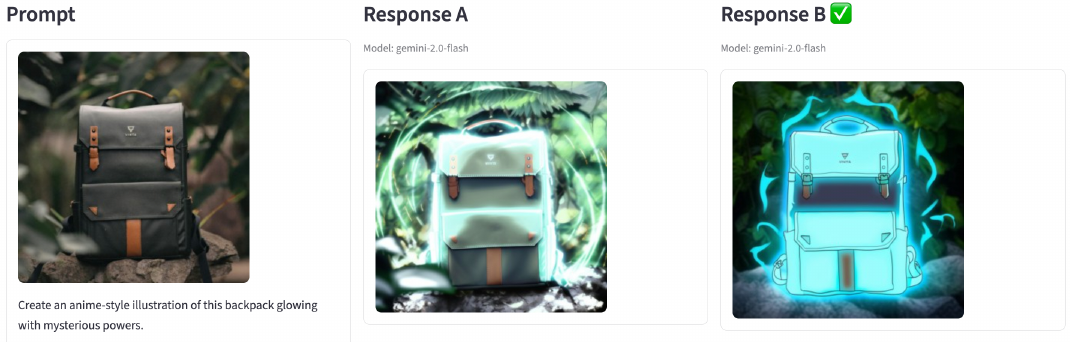}
    \caption{An example of MMRB2 image-editing task. Responses A and B are both generated by Gemini 2.0 Flash Image, while B is preferred over A. The rationale is that Response B follows the instruction better, and the backpack is more ``anime-styled.''}
    \label{fig:edit_ex2}
\end{figure}

\begin{figure}[h]
    \centering
    \includegraphics[width=\linewidth]{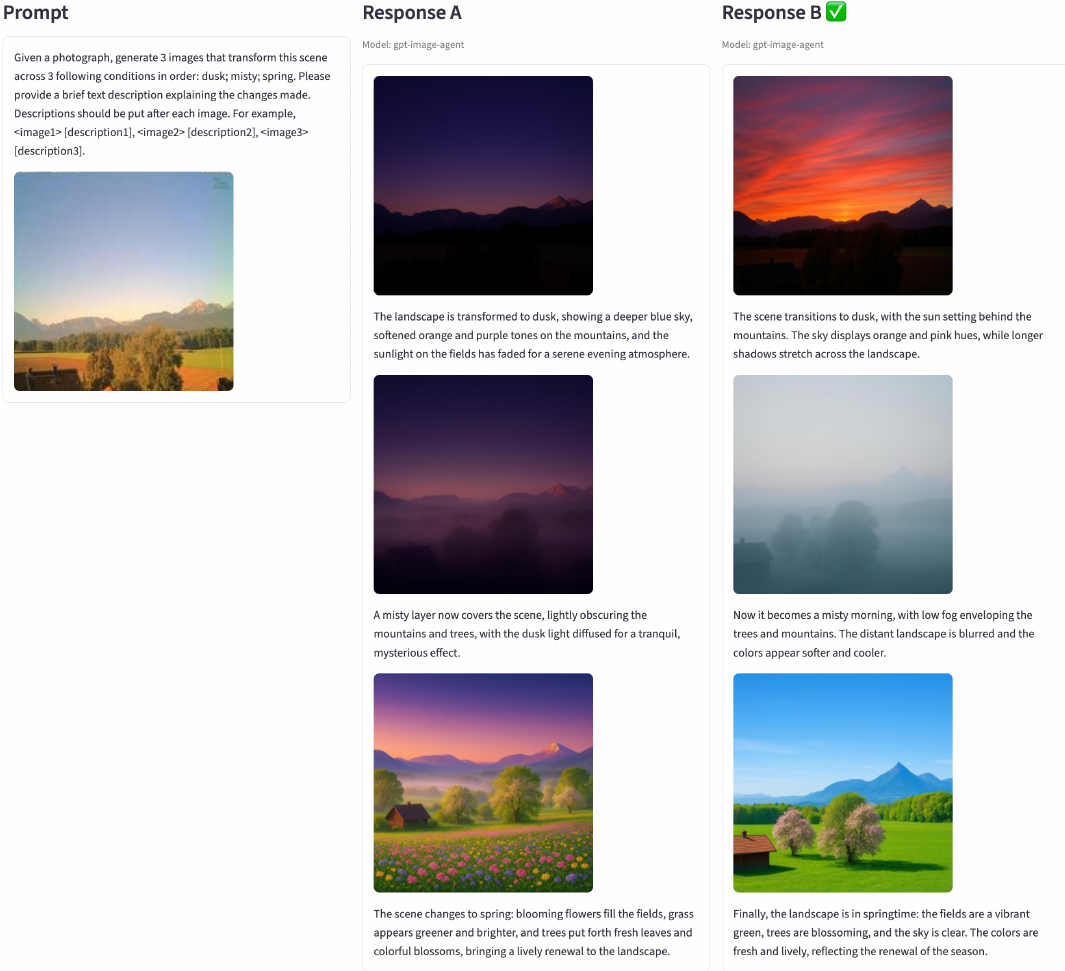}
    \caption{An example of MMRB2 interleaved task. Responses A and B are both generated by the agent with GPT-Image, while B is preferred over A. The rationale is that Response B better follows the instruction and is more consistent with the original image.}
    \label{fig:interleaved_ex1}
\end{figure}

\begin{figure}[h]
    \centering
    \includegraphics[width=\linewidth]{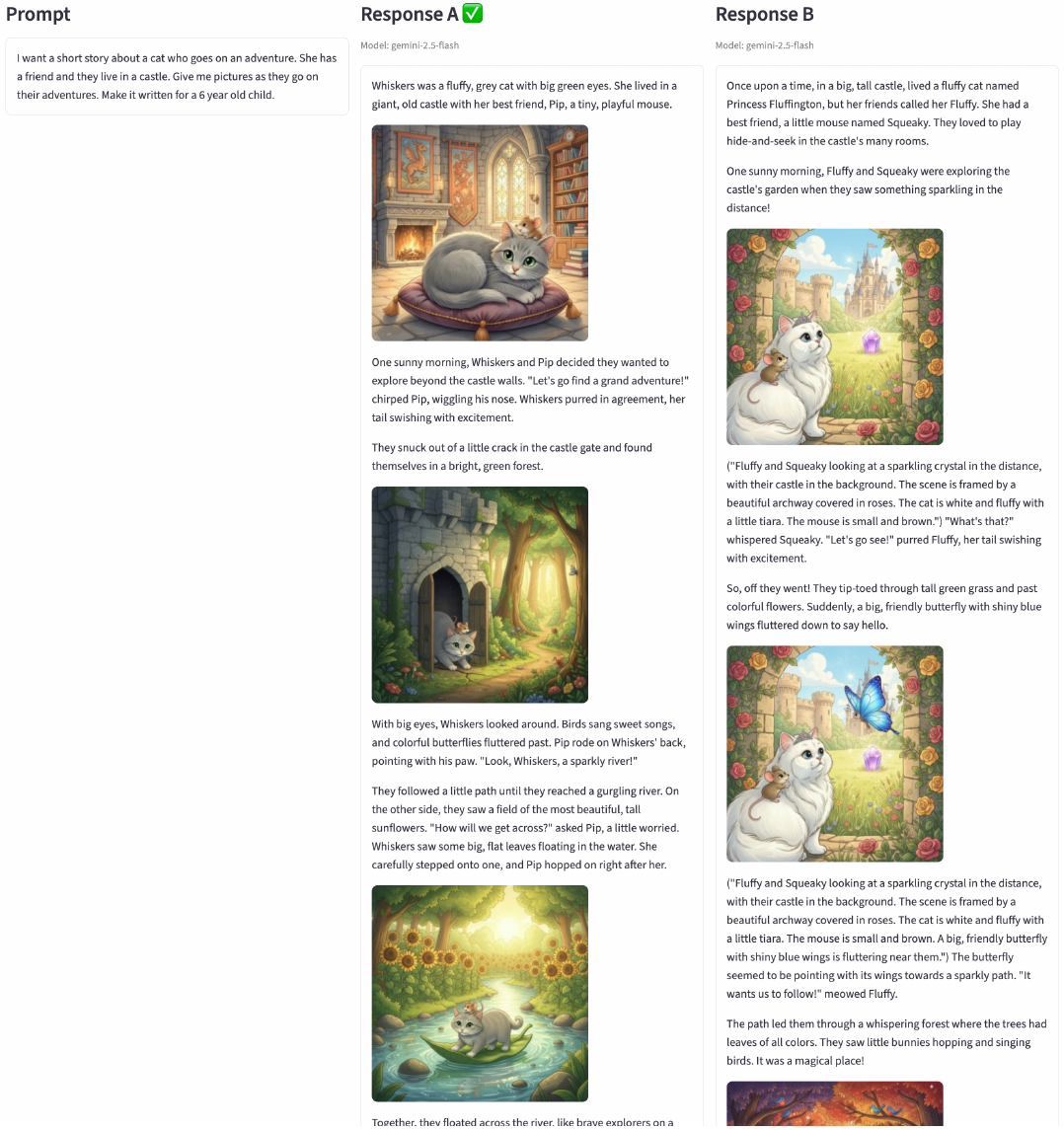}
\end{figure}

\begin{figure}[h]
    \centering
    \includegraphics[width=\linewidth]{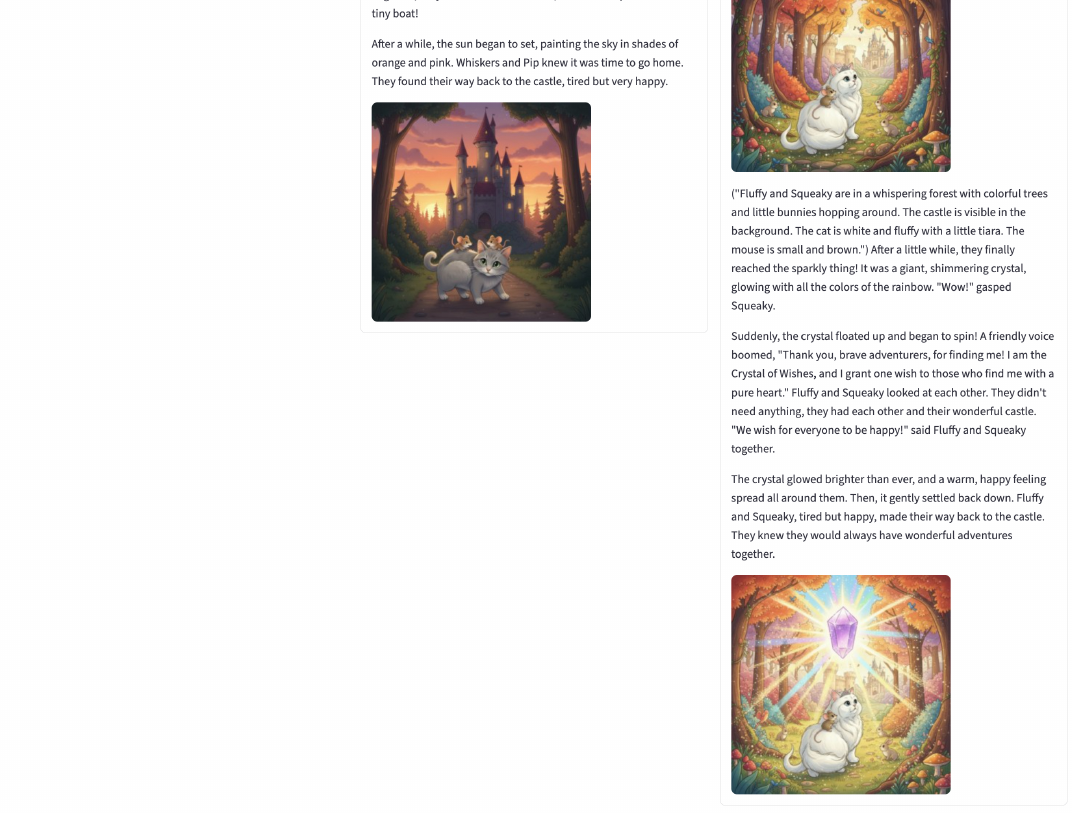}
    \caption{An example of MMRB2 interleaved task. Responses A and B are both generated by Gemini 2.5 Flash Image, while A is preferred over B. The rationale is that in Response B the cat is barely changed across the images, while in A the cats are more natural while remaining consistent.}
    \label{fig:interleaved_ex2}
\end{figure}

\begin{figure}[h]
    \centering
    \includegraphics[width=\linewidth]{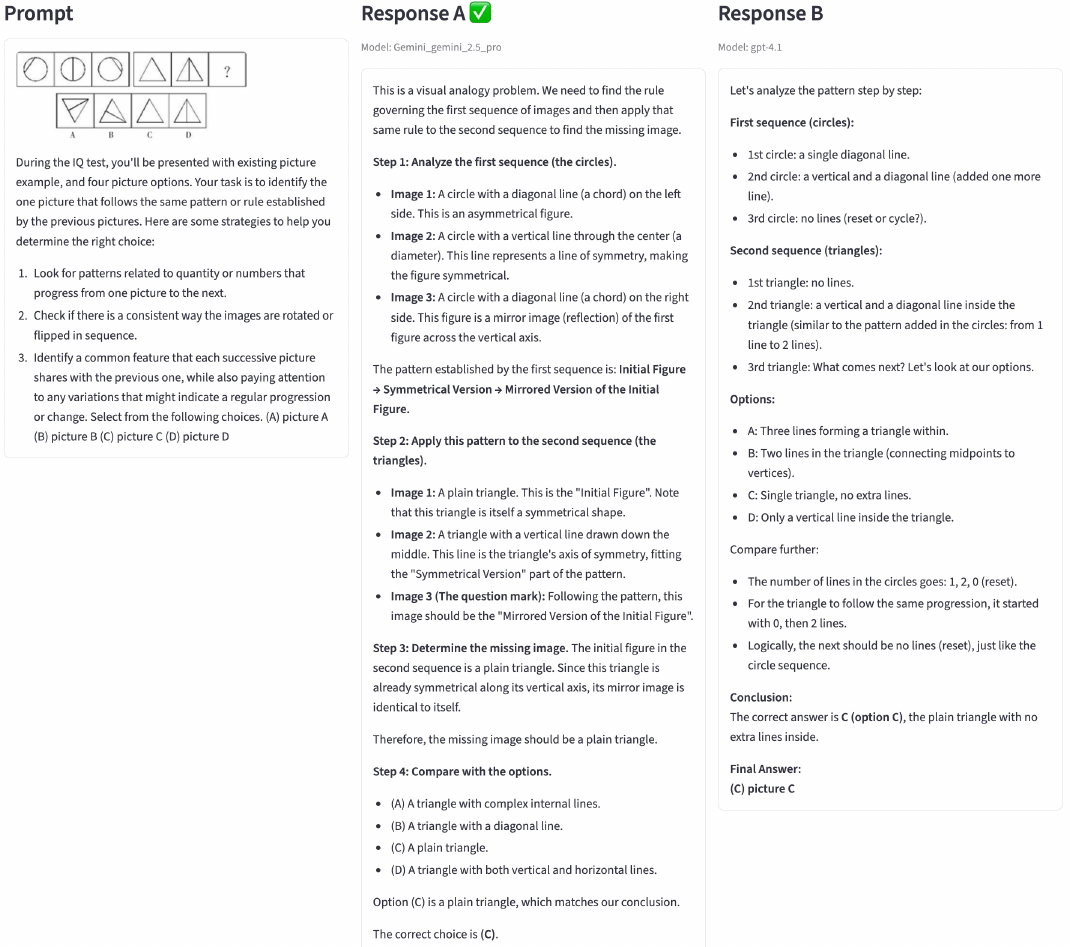}
    \caption{An example of MMRB2 multimodal reasoning task. Response A, generated by Gemini 2.5 Pro, is preferred over Response B, which is generated by GPT-4.1. Response A has correct reasoning and answer, while Response B's reasoning has apparent problems. For example, ``2nd circle: a veritcal and a diaglonal line'' is incorrect.}
    \label{fig:reasoning_ex1}
\end{figure}

\begin{figure}[h]
    \centering
    \includegraphics[width=\linewidth]{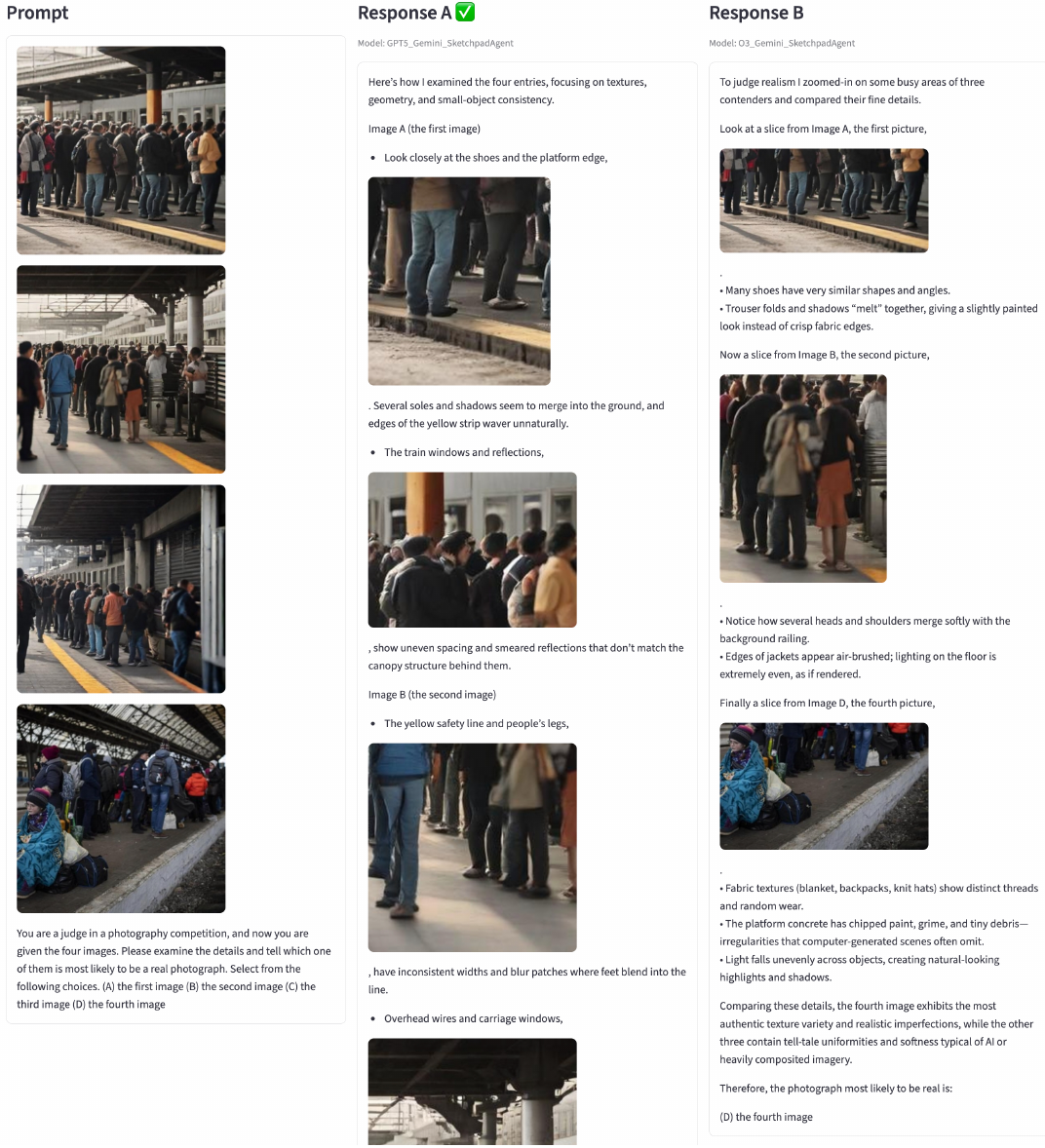}
\end{figure}

\begin{figure}[h]
    \centering
    \includegraphics[width=\linewidth]{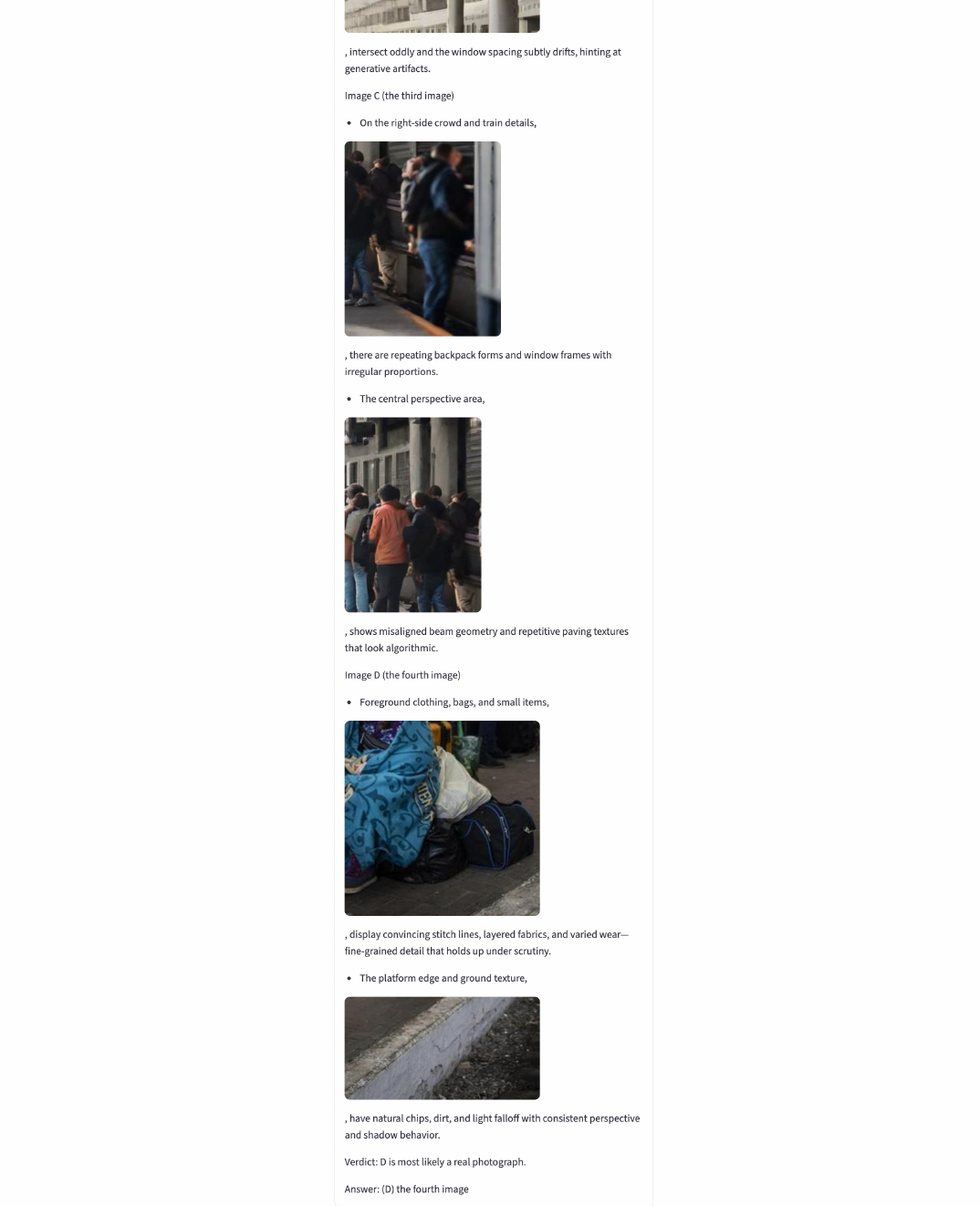}
    \caption{An example of MMRB2 multimodal reasoning task. Responses A and B are both generated by sketchpad agents. A uses GPT-5 as the LLM backbone, and B uses o3 as the backbone. A is preferred over B. The rationale is that B does not contain analysis for the third image, so the reasoning process is incomplete.}
    \label{fig:reasoning_ex2}
\end{figure}

\end{document}